\documentclass[10pt,journal,letterpaper,compsoc]{IEEEtran}
\usepackage{dsfont}
\usepackage{graphicx}
\usepackage{parskip}
\usepackage{epsfig}
\usepackage{epstopdf}
\usepackage{graphicx}
\usepackage{amsmath}
\usepackage{amsthm}
\usepackage{amssymb}
\usepackage{subfigure}
\usepackage{algorithm}
\usepackage{algorithmic}
\usepackage{color}
\usepackage{url}
\usepackage{multirow}
\usepackage{tikz}
\usepackage{textcomp}
\usepackage{cases}
\usepackage{makecell}
\usepackage{bm}
\usepackage{multirow}
\usepackage{bibspacing}
\usepackage{xcolor}
\usepackage{pifont} 

 \theoremstyle{plain}

 \theoremstyle{definition}

 \theoremstyle{remark}

\floatstyle{plain}
\newfloat{myalgo}{tbhp}{mya}

\ifCLASSOPTIONcompsoc
\else
\fi
\ifCLASSINFOpdf
\else
\fi
\begin{document}

\title{Wan-Alpha++: Animating a Single RGBA Asset into Controllable Transparent Video Effects}

\author{
Haotian Dong,
~Xin~Wang,
        Wenjing Wang,~\IEEEmembership{Member,~IEEE},
        Chen Li,
        Jing Lyu,
        Yan Tina Luximon,
        ~Ping~Li,~\IEEEmembership{Member,~IEEE},
        Wanchao Su,
        Ke Xu,
        ~Di~Lin,~\IEEEmembership{Member,~IEEE}

\IEEEcompsocitemizethanks{
\IEEEcompsocthanksitem Haotian Dong is with the College of Intelligence and Computing, Tianjin University, China.
\IEEEcompsocthanksitem Xin Wang is with the
Laboratory for Artificial Intelligence in Design
Limited (AiDLab), The Hong Kong Polytechnic University, Hong Kong.
\IEEEcompsocthanksitem Wenjing Wang is an independent researcher.
\IEEEcompsocthanksitem Chen Li is an independent researcher. 
\IEEEcompsocthanksitem Jing Lyu is an independent researcher. 
\IEEEcompsocthanksitem Yan Luximon is with the School
of Design, The Hong Kong Polytechnic University, Hong Kong.
\IEEEcompsocthanksitem Ping Li is with the Department of Computing, The Hong Kong Polytechnic University, Hong Kong (e-mail: lipingfire@gmail.com).
\IEEEcompsocthanksitem Wanchao Su is with the School of Software, Dalian University of Technology, China and SensiLab, Monash University, Australia.
\IEEEcompsocthanksitem Ke Xu is with the University of Science and Technology of China, China
\IEEEcompsocthanksitem Di Lin is with the College of Intelligence and Computing, Tianjin University, China.
}
}

\markboth{IEEE TRANSACTIONS ON PATTERN ANALYSIS AND MACHINE INTELLIGENCE}%
{Shell \MakeLowercase{\textit{et al.}}: Bare Demo of IEEEtran.cls for Computer Society Journals}

\IEEEcompsoctitleabstractindextext{%
\begin{abstract}

RGBA video generation is essential for creating compositable visual assets in filmmaking, game development, and digital content production.
However, generating high-quality RGB appearance alongside temporally coherent transparency remains a significant challenge. 
Existing text-to-RGBA video methods often entangle RGB and alpha distributions and offer limited control over the precise appearance of a target asset.
In addition, incorporating image conditioning typically suffers from the 
emergence-disappearance effects, which fail existing first-frame or first-and-last-frame conditions due to their fully transparent boundary frames.
To address these issues, we present Wan-Alpha++, a unified framework that brings controllable RGBA
video generation to both text- and image-to-video diffusion backbones through a shared RGBA latent
space, extending our conference version Wan-Alpha from the text-conditioned setting to
image-conditioned generation with an interior reference frame.
Our method first learns shiftable RGB-alpha distributions while preserving the pretrained RGB prior.
Specifically, we introduce a transparency-aware bidirectional diffusion objective to produce distinguishable RGBA latent representations, and a transparency-guided mean sampler that shifts the noise distribution to improve transparency generation and enable spatial controllability.
Building upon this RGBA generation prior, we introduce temporal phase reparameterization for image-conditioned effect generation. By cyclically
transforming a zero-positive-zero alpha pattern into a positive-zero-positive pattern, it relocates visible content to both temporal boundaries, facilitating effective appearance conditioning with limited RGBA data.
During inference, normalized transparent-state gradient guidance places a user-specified invisible interval, and an inverse phase transformation subsequently recovers the desired emergence-disappearance effect.
Extensive experiments demonstrate that Wan-Alpha++ achieves state-of-the-art performance in text-to-RGBA generation, precise spatial transparency control, and inference efficiency. For image-to-RGBA effect generation, Wan-Alpha++ significantly outperforms previous generation-and-matting pipelines, improving DINO-I from $0.6679$ to $0.9227$ and CLIP-I from $0.7969$ to $0.9368$,
while providing enhanced temporal consistency and controllable transparency evolution.

\end{abstract}



}

\maketitle

\IEEEdisplaynotcompsoctitleabstractindextext

\IEEEpeerreviewmaketitle

\section{Introduction}
RGBA extends conventional RGB representation with an alpha channel that describes per-pixel transparency, enabling foreground content to be directly composited with arbitrary backgrounds. As a result, RGBA videos are widely used in game development, filmmaking, television production, user-interface design, and other creative workflows. While automatically generating RGBA videos can substantially reduce manual effort, it faces unique challenges beyond standard RGB video generation: the model must jointly synthesize high-quality RGB appearance, spatially accurate transparency, and temporally coherent alpha evolution.


Early text-to-RGBA video generation methods~\cite{chen2025transanimate,TransVDM,layeranimate,ILDiff} extend RGBA image generation models to video frameworks. A representative strategy combines the 2D RGBA VAE of LayerDiffuse~\cite{LayerDiffuse} with pretrained video diffusion models such as AnimateDiff~\cite{AnimateDiff}. However, an image VAE does not explicitly model temporal consistency and tends to entangle RGB and alpha information in the latent space, resulting in degraded appearance, inaccurate transparency, and limited motion. Subsequent approaches improve temporal modeling through dedicated video architectures. TransPixeler~\cite{wang2025transpixeler}, for example, introduces alpha tokens, duplicates the generation backbone, and employs cross-RGBA attention for information exchange. Nevertheless, architectural interaction alone does not fully resolve the ambiguity between RGB appearance and alpha transparency; it also incurs substantial computational overhead and remains limited in synthesizing semi-transparent effects such as smoke, illumination, water, and translucent materials.

{To address these text-driven challenges, our preliminary conference work, Wan-Alpha~\cite{dong2026video}, explores RGB-alpha distribution learning for text-to-RGBA video generation. Instead of substantially modifying the pre-trained video-generation backbone, it preserves the learned RGB distribution while shifting the alpha distribution in both latent and noise spaces. Through a transparency-aware bidirectional diffusion objective that produces more distinguishable RGBA latent representations, a transparency-guided mean sampler that enhances transparency generation and enables spatial control, and a carefully constructed RGBA video dataset, Wan-Alpha can generate stable transparency without compromising the RGB generation capability of the pretrained model.}

{Nonetheless, text-to-RGBA video generation methods still fall short in achieving precise control over the visual appearance and spatial composition.}
{An RGBA reference image offers explicit appearance and transparency information, yet directly extending existing image-to-RGB models to RGBA video generation remains a challenging and largely unexplored task.
A fundamental limitation arises from their first-frame or first-and-last-frame conditioning mechanisms~\cite{wan2025wan,kong2024hunyuanvideo, cogvideox,bar2024lumiere, shi2024motion}, which assume that the target content is observable at temporal boundaries of the generated video. This assumption breaks down for RGBA effects, where complete disappearance of content at the first and last frames is a common and often desirable design choice.
A straightforward solution is to directly modify the condition mask to support an interior reference, which requires a large amount of data that is prohibitively expensive to collect.
Alternatively, inference-time gradient guidance may optimize a designated interior frame toward the reference image, but such a soft constraint cannot guarantee exact appearance and alpha preservation and may compromise generation quality.
Consequently, middle-frame-conditioned RGBA video generation faces two coupled challenges:  modality adaptation from RGB to RGBA generation and temporal-conditioning adaptation from boundary to interior reference frames.}


In this paper, we extend our Wan-Alpha~\cite{dong2026video} to address the above two challenges, by proposing Wan-Alpha++, a unified framework for controllable text-to-RGBA and data-efficient image-to-RGBA video generation.
Instead of directly modifying the conditioning pattern of the pretrained first-and-last-frame video model, we reparameterize the training distribution through visibility-aware temporal rotation.
Specifically, an RGBA effect that emerges from and subsequently vanishes into transparency is cyclically shifted such that its content-rich interior frames are moved to the temporal boundaries. The resulting sequences contain observable content at both endpoints, allowing us to retain the boundary-conditioning capability of the pretrained model while learning the distinctive temporal evolution of RGBA effects from limited data.
Based on the learned distribution, we introduce transparent-state guidance during inference: the same reference image is supplied at both boundaries, while a user-specified interior interval is guided toward a fully transparent target by minimizing a pixel-space MSE loss along the sampling trajectory.
Unlike an arbitrary RGB target, a fully transparent frame represents a low-complexity state and can therefore be reliably imposed through gradient guidance. Finally, we apply the inverse temporal rotation to recover the desired emergence-disappearance sequence. These two designs are mutually dependent: temporal rotation converts difficult interior conditioning into familiar boundary conditioning, while transparent-state guidance determines the invisible interval required to recover the target temporal pattern.



We summarize our major contributions below:
\begin{itemize}
  \item We propose Wan-Alpha, a high-quality text-to-RGBA video generation framework based on shiftable RGB–alpha distribution learning. It enables stable and controllable transparency generation while preserving the RGB synthesis capability of a pretrained video model.
  \item We introduce a transparency-aware bidirectional diffusion objective for distinguishable RGBA latent learning and a transparency-guided mean-shifting strategy for controllable RGBA generation. These designs improve joint RGB-alpha modeling in the latent and noise spaces, respectively.

  \item We construct a high-quality RGBA video dataset containing diverse motion, complex boundaries, and semi-transparent effects. Together with the proposed framework, it enables state-of-the-art performance across generation quality, temporal consistency, transparency rendering, and controllability.
  
  \item We propose Wan-Alpha++, which extends Wan-Alpha from text conditioning to image conditioning. By introducing temporal phase reparameterization to convert an interior reference into observable boundary conditions, and transparent-state gradient guidance to control the invisible interval and recover the desired emergence-disappearance effect, Wan-Alpha++ achieves middle-frame-conditioned RGBA video generation under limited training data.
\end{itemize}

\section{Related work}
\subsection{RGBA Image Generation}
Compared with RGB images~\cite{liopen,lin2022generative, wang2025hrc,dong2023cvsformer,duan2024twostage,IASD,csec,LinearColor}, high-quality RGBA images are substantially more difficult to acquire, making it expensive to train large-scale RGBA generative models from scratch. Existing approaches therefore commonly adapt pretrained RGB image models to transparency generation. LayerDiffuse~\cite{LayerDiffuse} introduces latent transparency by adapting the VAE of Stable Diffusion~\cite{latentdiffusion} to encode RGB and alpha information within a shared latent space, followed by parameter-efficient diffusion fine-tuning. AlphaVAE~\cite{alphavae} improves the RGBA representation and substantially reduces the amount of training data required for VAE adaptation. In contrast, Alfie~\cite{quattrini2024alfie} extracts transparency from the attention features of a pretrained diffusion model without additional training, but mainly supports transparent backgrounds and has limited ability to represent semi-transparent content. Subsequent studies extend RGBA generation to multi-layer image synthesis~\cite{DreamLayer,yang2025generative,huang2025psdiffusion,kang2025layeringdiff} and transparent image inpainting~\cite{dai2025trans}. 

Although these methods demonstrate the potential of pretrained RGB priors for transparency generation, they operate on individual images and do not explicitly model temporal alpha evolution.

\subsection{RGBA Video Generation}
RGBA video generation additionally requires temporally coherent motion and transparency, while suitable training data remain scarce. Early approaches~\cite{chen2025transanimate,TransVDM,layeranimate,ILDiff} combine the image-based LayerDiffuse VAE~\cite{LayerDiffuse} with video generation frameworks such as AnimateDiff~\cite{AnimateDiff}, but the lack of explicit temporal modeling in the image VAE often results in unstable alpha channels and limited motion. Other studies develop dedicated architectures for transparent or layered video generation. LayerFlow~\cite{ji2025layerflow} formulates layer-aware video synthesis within a unified generation framework, while TransPixeler~\cite{wang2025transpixeler} duplicates the video backbone and introduces cross-RGBA attention to exchange information between RGB and alpha branches. 

However, these methods either incur substantial computational overhead or remain limited in distinguishing the distribution of RGB appearance from that of transparency. Our preliminary conference work Wan-Alpha~\cite{dong2026video} learns shiftable RGB–alpha distributions in the latent and noise spaces, enabling stable and controllable text-to-RGBA video generation while preserving the RGB synthesis capability of the pretrained model. Despite being able to 
generate high-quality and diverse videos, text-to-RGBA video generators typically offer less satisfactory spatial controllability ({\it e.g.}, animating the appearance and alpha structure of a user-provided asset).

\subsection{Image-to-RGB Video Generation}
{We review related Image-to-RGB video generation methods, as Wan-Alpha++ is the first Image-to-RGBA video generation model.

\noindent{\bf Boundary-Frame-Conditioned Generation~~}
Recent image-to-video {generation (I2V)} methods generally use reference image(s) to constrain visual appearance while synthesizing 
temporal dynamics. A common paradigm places the reference at the first frame. Stable Video Diffusion~\cite{blattmann2023stable} adapts 
latent video diffusion models to animate a given image.
DynamiCrafter~\cite{xing2024dynamicrafter} incorporates both latent-level image features and pixel-level image information into a pretrained text-to-video model. CogVideoX-I2V~\cite{cogvideox} encodes the input image with a 3D VAE, pads its latent representation along the temporal dimension, and concatenates it with the noisy video latent. Wan~\cite{wan2025wan} further scales this paradigm to large diffusion transformers. FrameBridge~\cite{wang2024framebridge} reformulates {I2V}
as a data-to-data bridge process, treating the input image as the starting point of generation instead of an auxiliary condition.

Another line of work incorporates both first- and last-frame conditions to constrain video endpoints. PixelDance~\cite{zeng2024make} uses a strict first-frame condition together with an optional last-frame instruction to generate motion-rich videos. SEINE~\cite{chen2023seine} adopts random-mask training and supports first-and-last-frame transition as one of several frame-completion configurations. ToonCrafter~\cite{xing2024tooncrafter} specializes in dual-frame conditioning for generative cartoon interpolation, while ViBiDSampler~\cite{yang2025vibidsampler} performs training-free bidirectional sampling by alternately conditioning forward and backward generation on the two endpoints. Wan2.1-FLF2V~\cite{wan2025wan} directly trains a large video model with first-and-last-frame conditions.

These methods demonstrate the effectiveness of boundary images for appearance and endpoint control. However, their conditioning mechanisms are tied to predefined temporal positions and are unsuitable when the informative reference corresponds to an interior frame, as frequently occurs in emergence–disappearance RGBA effects.

\noindent{\bf Training-Free Guidance for Video Generation~~}
Training-free diffusion guidance provides an alternative {by controlling pretrained models without fine-tuning on specific tasks.}
FreeDoM~\cite{yu2023freedom} and MPGD~\cite{he2024manifold} construct differentiable objectives to guide diffusion sampling while preserving the pretrained generative prior. Recent studies extend this paradigm to video generation. {Specifically, }MOFT~\cite{xiao2024video} extracts motion-aware features from video diffusion models and optimizes noisy latents using feature-correlation losses for training-free motion control. TITAN-Guide~\cite{simon2025titan} further introduces memory-efficient forward-gradient optimization for text-to-video diffusion models and demonstrates its applicability to frame interpolation. FlexTI2V~\cite{lai2025unified} supports visual conditions at arbitrary temporal positions through dynamic latent-patch exchange, although it does not explicitly use gradient guidance.

These approaches improve conditioning flexibility without model retraining, but generally impose soft sampling-time constraints and cannot guarantee exact reproduction of a target frame.


\section{Method}
\subsection{Overview}
Wan-Alpha++ aims to generate an RGBA video from a user-provided RGBA image while {faithfully} preserving its appearance and {enabling}
controllable emergence {and} 
disappearance process. 
{Building} upon our preliminary text-to-RGBA video generation framework~\cite{dong2026video}, {Wan-Alpha++ substantially advances the previous version and}
learns distinguishable RGB-alpha representations and controllable transparency distributions. We first revisit the RGBA VAE training architecture in Sec. 3.2, separating RGB and alpha information in {a}
latent space. Based on this RGBA representation prior, we construct an image-conditioned model in Sec. 3.3. We first learn a controllable spatial distribution by introducing a transparency-guided sampler. Directly learning middle-frame conditioning requires adapting the pretrained
image-to-video model to a temporal conditioning pattern unseen during
pretraining, which typically demands substantial training data. This is
particularly costly for RGBA video generation because high-quality RGBA videos
are scarce and expensive to collect. We instead reparameterize the temporal phase of RGBA effects, converting a sequence whose informative content lies in the interior into one with observable boundary frames.
This allows the model to retain the pretrained first-and-last-frame conditioning mechanism. During inference, transparent-state guidance determines an interior invisible interval, after which an inverse temporal transformation recovers the desired emergence-disappearance effect. In Sec. 3.4, we introduce the construction of our RGBA dataset for VAE and DiT. All components enable our Wan-Alpha++ to achieve state-of-the-art performance across multiple metrics.

\subsection{{Shiftable RGBA Distribution Learning}}
\begin{figure*}[t]
    \centering
    \includegraphics[width=0.99\linewidth]
    {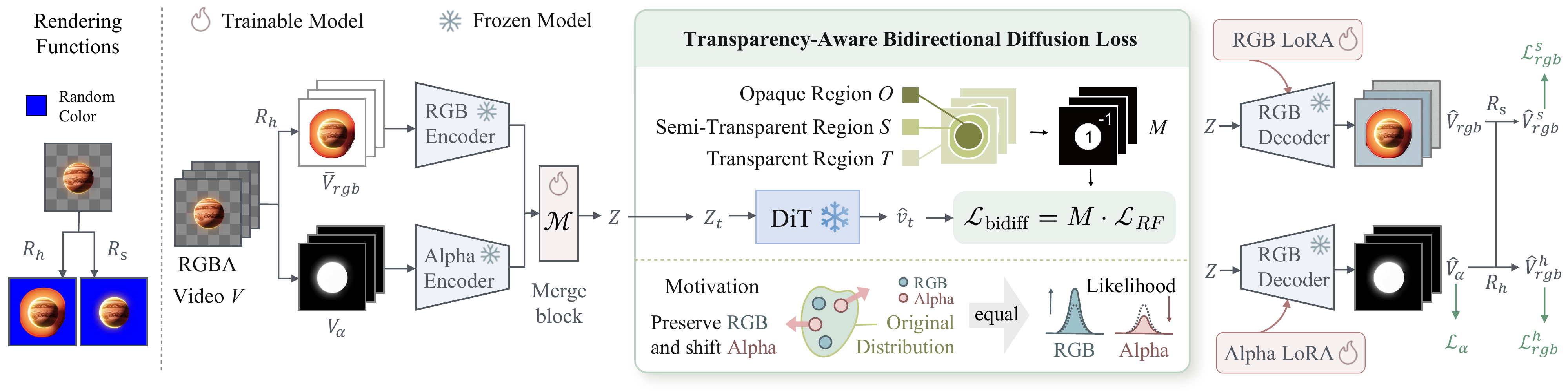}
    \vspace{-2mm}
    \caption{Training framework of our RGBA VAE. The RGB and alpha channels are
    encoded separately and fused into a shared latent representation. The
    transparency-aware bidirectional diffusion objective preserves the
    pretrained RGB distribution while shifting alpha-related representations.
    Rendering-based reconstruction objectives further supervise RGB appearance
    and alpha transparency.}
    \label{fig:rgba_vae_framework}
\end{figure*}
\subsubsection{{Distinguishable RGBA Latent Representation}}
\label{wanvae}
{As illustrated in Figure~\ref{fig:rgba_vae_framework}, we start from a pretrained RGB video generation model~\cite{wan2025wan} consisting of a video VAE and a diffusion transformer. Given an RGBA video, {we denote the video {$V$ as a {
combination of} RGB component}
$V_{rgb}$ and {transparency component} $V_\alpha$, {i.e.,} $V = (V_{rgb}, V_\alpha)$.
We aim to embed RGB appearance and alpha transparency into a shared latent representation without changing its original dimensionality. This {enables}
the pretrained video diffusion model to be adapted {for}
RGBA generation without {requiring any modifications to}
its backbone architecture.}

{Directly encoding $V_{rgb}$ and $V_\alpha$ into the same latent space tends to entangle background color with transparency. We therefore
introduce {a} hard and {a} soft rendering function:
\begin{equation}
\label{eq:soft_render}
\left\{
\begin{aligned}
    R_h(V_{\mathrm{rgb}}, V_{\alpha}, c)
    &=\!
    V_{\mathrm{rgb}} \odot \mathbb{I}(V_{\alpha} > 0)
    \!+\! c \!\odot\! \left[1\!-\!\mathbb{I}(V_{\alpha} > 0)\right],
    \\
    R_s(V_{\mathrm{rgb}}, V_{\alpha}, c)
    &=\!
    V_{\mathrm{rgb}} \odot V_{\alpha}
    + \!c\! \odot\! (1-V_{\alpha}),
\end{aligned}
\right.
\end{equation}
where $\odot$ denotes element-wise multiplication, $\mathbb{I}(\cdot)$ is an
indicator function, and $c$ is {a} randomly sampled {color value} from a predefined color set C=$\{$black, blue, green, cyan, red, magenta, yellow, white$\}$.
}

{
The hard-rendered RGB video and the three-channel replicated alpha video are
separately encoded by {a}
frozen video encoder $E$. A trainable feature merge block $\mathcal{M}$ then produces the shared RGBA latent:
\begin{equation}
    Z =
    \mathcal{M}\left(
        E\left(R_h(V_{\mathrm{rgb}},V_{\alpha},c)\right),
        E(V_{\alpha})
    \right).
    \label{eq:rgba_latent}
\end{equation}
The latent $Z$ is decoded by two branches equipped with RGB and alpha LoRA parameters:
$\widehat V_{\mathrm{rgb}} = D_{\mathrm{rgb}}(Z)$ and $
    \widehat V_{\alpha} = D_{\alpha}(Z)$.

{Given a reconstruction $\widehat X$ and its target $X$, we formulate pixel, perceptual, and edge-consistency objectives to supervise reconstruction under different transparency states as:
}
\begin{equation}
\label{eq:reconstruction_loss}
\left\{
\begin{aligned}
    \mathcal{L}_{\mathrm{pix}}(\widehat X,X)
    &= \|\widehat X-X\|_1,
    \\
    \mathcal{L}_{\mathrm{perc}}(\widehat X,X)
    &= \|\Phi(\widehat X)-\Phi(X)\|_2^2,
    \\
    \mathcal{L}_{\mathrm{edge}}(\widehat X,X)
    &= \|S(\widehat X)-S(X)\|_1,
    \\
    \mathcal{L}_{\mathrm{rec}}(\widehat X,X)
    &=
    \mathcal{L}_{\mathrm{pix}}
    + \lambda_{\mathrm{perc}}\mathcal{L}_{\mathrm{perc}}
    + \lambda_{\mathrm{edge}}\mathcal{L}_{\mathrm{edge}},
\end{aligned}
\right.
\end{equation}
where $\Phi$ is a pretrained perceptual feature extractor, $S$ denotes the
Sobel operator, {and $\mathcal{L}_{\mathrm{rec}}$ is the overall reconstruction objective.}} {Using independently sampled colors $c_s$ and $c_h$, {the reconstructed rendered videos $\widehat V_{\mathrm{rgb}}^{s}$ and $\widehat V_{\mathrm{rgb}}^{h}$, and the corresponding ground-truth rendered videos $V_{\mathrm{rgb}}^{s}$ and $V_{\mathrm{rgb}}^{h}$ are formulated as:}
\begin{equation}
\left\{
    \begin{aligned}
    &\widehat V_{\mathrm{rgb}}^{s}
    =\! R_s(\widehat V_{\mathrm{rgb}},\widehat V_{\alpha},c_s),~
    V_{\mathrm{rgb}}^{s}
    =\! R_s(V_{\mathrm{rgb}},V_{\alpha},c_s),
    \\
    &\widehat V_{\mathrm{rgb}}^{h}
    =\! R_h(\widehat V_{\mathrm{rgb}},\widehat V_{\alpha},c_h),~
    V_{\mathrm{rgb}}^{h}
    =\! R_h(V_{\mathrm{rgb}},V_{\alpha},c_h).
    \end{aligned}
\right.
\end{equation}
The rendering objective {$\mathcal{L}_{\mathrm{render}}$} is then formulated as:
\begin{equation}
\label{eq:rendering_objective}
\left\{
\begin{aligned}
    &\mathcal{L}_{\alpha}
    =
    \mathcal{L}_{\mathrm{rec}}
    (\widehat V_{\alpha},V_{\alpha}),~
    \mathcal{L}_{\mathrm{rgb}}^{s}
    =
    \mathcal{L}_{\mathrm{rec}}
    (\widehat V_{\mathrm{rgb}}^{s},V_{\mathrm{rgb}}^{s}),
    \\
    &\mathcal{L}_{\mathrm{rgb}}^{h}\!
    =\!
    \mathcal{L}_{\mathrm{rec}}
    (\widehat V_{\mathrm{rgb}}^{h},V_{\mathrm{rgb}}^{h}),\!
    \mathcal{L}_{\mathrm{render}}
    =\!
    \mathcal{L}_{\alpha}\!
    +\! \mathcal{L}_{\mathrm{rgb}}^{s}
    +\! \mathcal{L}_{\mathrm{rgb}}^{h}.
\end{aligned}
\right.
\end{equation}
}\noindent{\bf {Transparency-Aware Bidirectional Diffusion Objective}~~}
{Preserving the original RGB latent distribution maintains the generative
capability of the pretrained DiT, whereas shifting alpha-related latents makes
transparency more distinguishable. Measuring only statistical distances between
latent features, however, does not guarantee that these representations are
distinguishable to the video DiT. We therefore introduce the frozen DiT into VAE
training and optimize the latent representation from its perspective.}

{
Let $t\in[0,1]$ denote the rectified-flow timestep. The noisy latent and target
velocity are:
\begin{equation}
    Z_t = (1-t)Z+t\epsilon,
    \quad
    v_t = \epsilon-Z,
    \quad
    \epsilon\sim\mathcal{N}(0,I).
    \label{eq:rectified_flow}
\end{equation}
Given the velocity $\widehat v_t$ predicted by the frozen DiT, the point-wise
rectified-flow loss is:
\begin{equation}
    \ell_{\mathrm{RF}}(p)
    =
    \left\|
        \widehat v_t(p)-v_t(p)
    \right\|_2^2,
    \label{eq:pointwise_rf}
\end{equation}
where $p$ indexes a 
spatial-temporal {location in the latent space.} Aggregating the point-wise errors over all spatial-temporal latent locations
gives the standard rectified-flow objective:
\begin{equation}
    \mathcal{L}_{\mathrm{RF}}
    =
    \frac{1}{|\Omega|}
    \sum_{p\in\Omega}
    \ell_{\mathrm{RF}}(p)
    =
    \frac{1}{|\Omega|}
    \sum_{p\in\Omega}
    \left\|
        \widehat v_t(p)-v_t(p)
    \right\|_2^2,
    \label{eq:global_rf}
\end{equation}
where $\Omega$ denotes the set of all spatial-temporal locations in the latent
representation. This standard objective assigns equal importance to opaque,
semi-transparent, and fully transparent regions.
}

{
We resize the alpha video to the latent resolution and define a signed mask:
\begin{equation}
    M(p)
    =
    \begin{cases}
        1, & p\in\mathcal{O},\\
        -1, & p\in\mathcal{S}\cup\mathcal{T},
    \end{cases}
    \label{eq:bidirectional_mask}
\end{equation}
where $\mathcal{O}$, $\mathcal{S}$, and $\mathcal{T}$ denote opaque,
semi-transparent, and fully transparent regions, respectively. The
transparency-aware bidirectional diffusion objective is:
\begin{equation}
    \mathcal{L}_{\mathrm{bidiff}}
    =
   M\cdot \mathcal{L}_{RF}.
    \label{eq:bidirectional_loss}
\end{equation}
Positive gradients preserve high likelihood for opaque RGB content, whereas
reversed gradients shift non-opaque representations away from the original RGB
distribution. The complete VAE objective is:
\begin{equation}
    \mathcal{L}_{\mathrm{VAE}}
    =
    \mathcal{L}_{\mathrm{render}}
    + 
      \mathcal{L}_{\mathrm{bidiff}}.
    \label{eq:vae_total}
\end{equation}
}
\subsection{Controllable Image-to-RGBA Generation}
\label{sec:phase_i2v}
In Figure~\ref{fig:image_rgba_training_framework}, we introduce the training framework of our RGBA video generator. Figure~\ref{fig:sampler} further illustrates the transparency-guided noise distribution. Based on the trained DiT, Figure~\ref{fig:image_rgba_framework} presents the process of transparent-state gradient guidance during inference.

\subsubsection{Boundary-Conditioned RGBA Initialization}

\label{sec:boundary_initialization}
We first transfer the learned RGBA representation to a pretrained
image-to-video model. Given a reference image, its encoded latent is placed
at the first temporal position, while the remaining positions are zero-padded.
A binary temporal mask indicates the position of the image condition. We
initially train on generic RGBA videos to obtain an image-to-RGBA video generation model and
subsequently extend the same conditioning interface to the
first-and-last-frame setting.

Given an RGBA video $V=\{V_i\}_{i=0}^{T-1}$, the first-and-last-frame condition
is represented as:
\begin{equation}
    C_i
    =
    \begin{cases}
        E_{\mathrm{RGBA}}(V_0), & i=0,\\
        E_{\mathrm{RGBA}}(V_{T-1}), & i=T-1,\\
        0, & \text{otherwise},
    \end{cases}
    \label{eq:flf_condition}
\end{equation}
with the corresponding temporal mask:
\begin{equation}
    B_i
    =
    \mathbb{I}\left(i\in\{0,T-1\}\right).
    \label{eq:flf_mask}
\end{equation}
The condition latent $C$, condition mask $B$, and noisy video latent are jointly
provided to the denoising model. This initialization transfers the
appearance-conditioning capability of the RGB video foundation model to the
RGBA domain, but it still assumes that informative content {exists}
at the
temporal boundaries.

\subsubsection{Visibility-Aware Temporal Phase Reparameterization}
\label{sec:phase_reparameterization}
Many RGBA effects follow an emergence-disappearance pattern. As illustrated in Figure~\ref{fig:image_rgba_training_framework}, they start
from a fully transparent state, gradually become visible, and eventually {fade out.}
For frame index $i$, we define {a}
temporal alpha mass {$m$} as:
\begin{equation}
    m_i
    =
    \frac{1}{HW}
    \sum_{x=1}^{W}
    \sum_{y=1}^{H}
    V_{\alpha,i}(x,y).
    \label{eq:alpha_mass}
\end{equation}
An emergence-disappearance effect approximately satisfies:
\begin{equation}
    m_0\approx 0,
    \qquad
    \max_i m_i>0,
    \qquad
    m_{T-1}\approx 0.
    \label{eq:zero_positive_zero}
\end{equation}
Its original zero-positive-zero pattern therefore contains no effective
appearance information at either temporal boundary.

Instead of moving the image condition to an interior position and retraining the
conditioning interface, we move the temporal origin of the video. {We design a cyclic temporal transformation $\mathcal{R}_{k}$ to reparameterizes $V$ into a phase-shifted sequence $\widetilde V$ by setting its $k$-th frame as the new temporal origin:
\begin{equation}
    \widetilde V
    =
    \mathcal{R}_{k}(V),~~
    \widetilde V_i
    =
    V_{(i+k)\bmod T},~~
    i=0,\ldots,T-1,
    \label{eq:cyclic_rotation}
\end{equation}
To ensure that both transformed boundaries contain observable content, the
cutting position {$k$} is sampled from:
\begin{equation}
    k\sim\mathcal{K},~~
    \mathcal{K}
    =
    \left\{
        j\;\middle|\;
        m_j>\tau_m,\;
        m_{(j-1)\bmod T}>\tau_m
    \right\},
    \label{eq:visible_cut}
\end{equation}
where $\tau_m$ is a visibility threshold. Consequently, {the existence pattern becomes:}
\begin{equation}
    0\rightarrow+\rightarrow0
    \quad
    \xrightarrow{\mathcal{R}_k}
    \quad
    +\rightarrow0\rightarrow+.
    \label{eq:phase_pattern}
\end{equation}}

The transformation changes only the temporal phase and preserves the frames,
motion, and transparency evolution of the original effect. We use
$\widetilde V_0$ and $\widetilde V_{T-1}$ as image conditions and fine-tune the
RGBA FLF2V model on $\widetilde V$. Sampling $k$ from $\mathcal{K}$ introduces
phase diversity while maintaining valid boundary conditions. This strategy
converts the {challenging}
middle-frame conditioning problem into {a simpler}
boundary-conditioning pattern already supported by the pretrained model.

\subsubsection{{Transparency-Guided Noise Distribution}}
\label{sec:mean_sampler}
\begin{figure*}[t]
    \centering
    \includegraphics[width=0.99\linewidth]
    {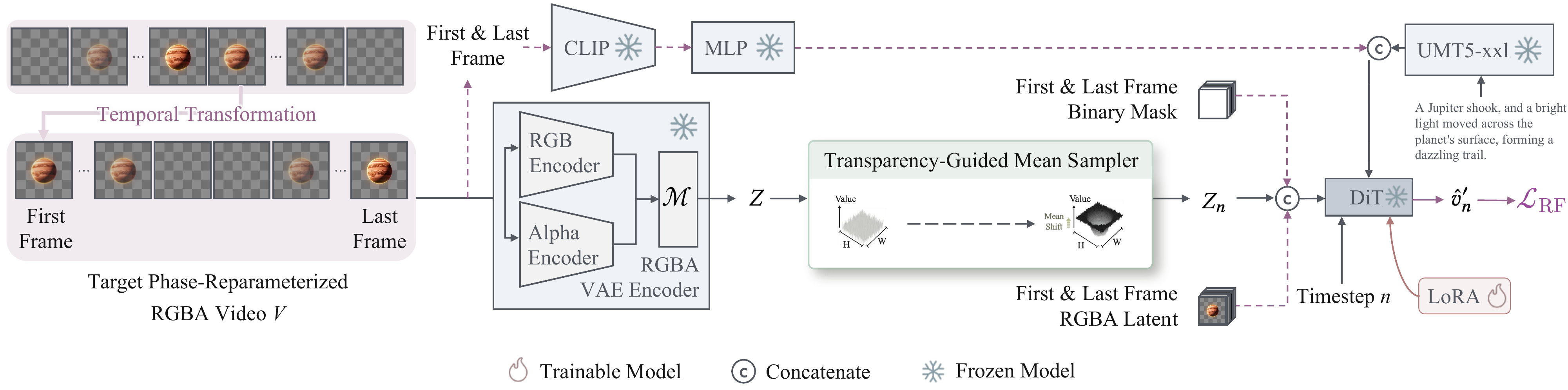}
    \vspace{-2mm}
    \caption{Training framework of our RGBA video generator. The
    pretrained RGBA VAE encodes the input video into the shared latent space.
    Transparency-guided mean sampler
    provides a transparency prior for DiT adaptation. }
    \label{fig:image_rgba_training_framework}
\end{figure*}

\begin{figure}[t]
    \centering
    \includegraphics[width=0.99\linewidth]
    {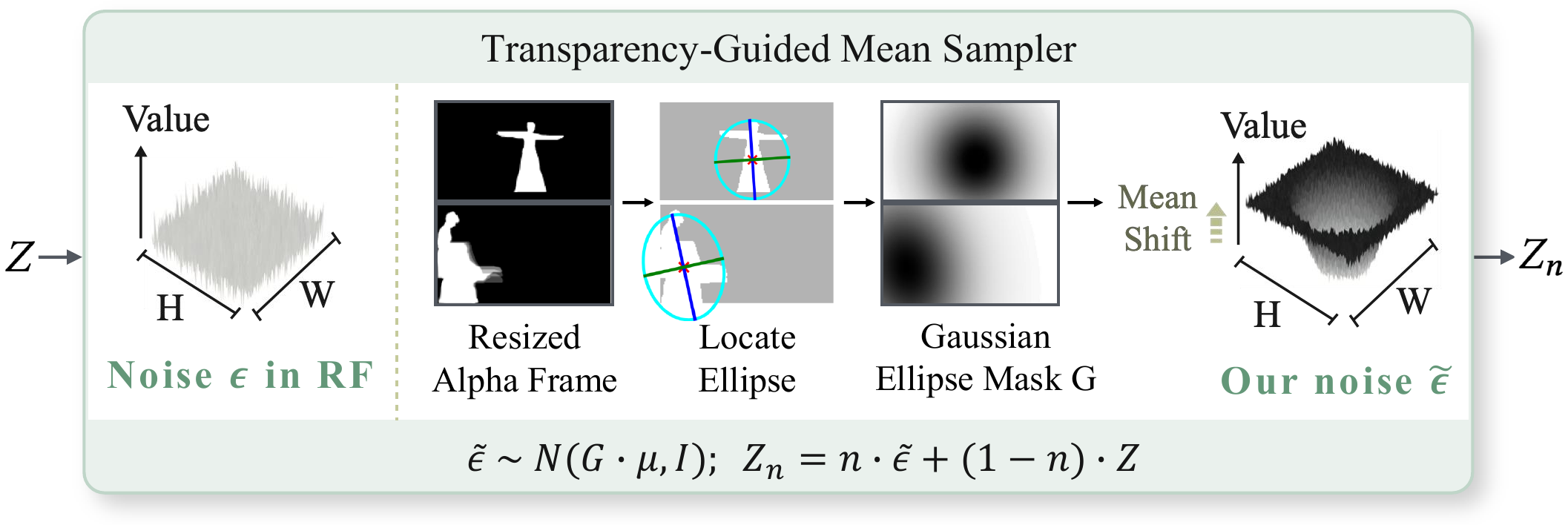}
    \vspace{-4mm}
    \caption{
    A Gaussian ellipse derived from the alpha channel shifts the noise mean,
    providing a transparency prior for DiT adaptation. During inference, the
    mask can be specified by users to control the approximate position and
    extent of generated content.}
    \label{fig:sampler}
\end{figure}
{
As illustrated in Figure~\ref{fig:image_rgba_training_framework}, after learning the shared RGBA representation, we adapt the DiT to RGBA video
generation. In Figure~\ref{fig:sampler}, we further introduce the transparency-guided noise distribution in detail. Standard rectified flow samples noise from a zero-mean Gaussian
distribution and therefore provides no prior regarding the spatial arrangement
of transparency. We instead shift the noise mean using a spatial mask derived
from the alpha channel:
    $\widetilde\epsilon
    \sim
    \mathcal{N}(G\cdot\mu,I)$,
where $G\in[0,1]^{T'\times H'\times W'}$ is a Gaussian ellipse mask and $\mu$
controls the mean-shift strength.
}

{
To construct $G$, we resize and binarize each alpha frame $A_i$ {and collect its positive-support coordinates $\mathcal{P}_i$, which is formulated as:}
\begin{equation}
\left\{
    \begin{aligned}
    B_i(x,y)=\mathbb{I}\left(A_i(x,y)>\tau_{\alpha}\right), \\
    \mathcal{P}_i
    =
    \left\{
        (x,y)\mid B_i(x,y)=1
    \right\},
    \end{aligned}
\right.
\end{equation}
where $\tau_{\alpha}$ is set as 0.5.
{The coordinate mean $\bm{m}$ and covariance $\Sigma$ are formulated as}:
\begin{equation}
\left\{
\begin{aligned}
    \bm{m}_i
    &=
    \frac{1}{|\mathcal{P}_i|}
    \sum_{\bm{q}\in\mathcal{P}_i}\bm{q},
    \\
    \Sigma_i
    &=
    \frac{1}{|\mathcal{P}_i|-1}
    \sum_{\bm{q}\in\mathcal{P}_i}
    (\bm{q}-\bm{m}_i)(\bm{q}-\bm{m}_i)^{\top}.
\end{aligned}
\right.
\end{equation}

Eigenvalue decomposition
$\Sigma_i\!=\!U_i\Lambda_iU_i^{\top}$ provides the principal directions and
orientation of the fitted ellipse. For coordinates $(x_i',y_i')$ expressed in
the principal-axis coordinate system, the Gaussian ellipse {$G$ is formulated as}:
    $G_i(x,y)
    =
    \exp\left[
        -\frac{1}{2}
        \left(
            \frac{{x_i'}^2}{(a_i/2)^2}
            +
            \frac{{y_i'}^2}{(b_i/2)^2}
        \right)
    \right]$,
where $a_i$ and $b_i$ are the lengths of the major and minor axes.
}

{
The shifted rectified-flow path and target velocity are:
\begin{equation}
    Z_n=(1-n)Z+n\widetilde\epsilon,
    \qquad
    \widetilde v_n=\widetilde\epsilon-Z.
    \label{eq:shifted_rf}
\end{equation}
The DiT is optimized with
  $  \mathcal{L}_{\mathrm{DiT}}
    =
    \left\|
        \widehat v_n-\widetilde v_n
    \right\|_2^2$.
During inference, $G$ can be specified by the user to control the approximate
location and extent of generated content.
}

\subsubsection{Transparent-State Gradient Guidance}
\label{sec:transparent_guidance}

\begin{figure*}[t]
    \centering
    \includegraphics[width=\linewidth]
    {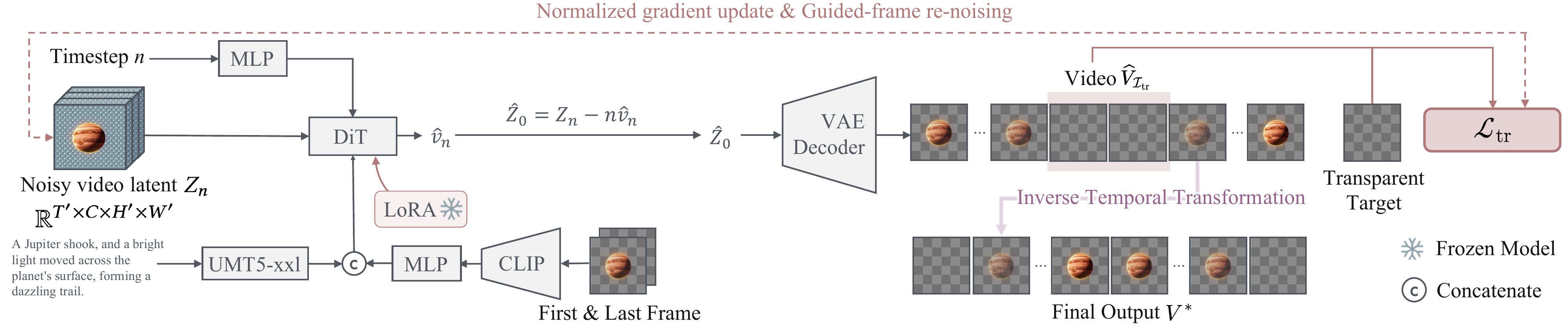}
    \vspace{-2mm}
    \caption{Inference framework of the proposed
    phase-reparameterized image-to-RGBA video generation.
    The same input RGBA image conditions both temporal boundaries. Transparent-state guidance applies normalized gradient updates
    to a user-specified invisible interval. Finally, temporal transformation
    recovers the desired emergence-disappearance effect.}
    \label{fig:image_rgba_framework}
\end{figure*}

After phase reparameterization, the model learns the
positive-zero-positive temporal pattern.
As illustrated in Figure~\ref{fig:image_rgba_framework}, during inference, we place the same
user-provided RGBA image $I_{\mathrm{ref}}$ at both temporal boundaries {$C_0$ and $C_{T-1}$, which is formulated as}:
\begin{equation}
    C_0=C_{T-1}=E_{\mathrm{RGBA}}(I_{\mathrm{ref}}).
    \label{eq:duplicate_reference}
\end{equation}
These duplicated conditions anchor the appearance and spatial alpha structure
of the generated effect. The user further specifies an interior interval {$\mathcal{I}_{\mathrm{tr}}$ as}:
\begin{equation}
    \mathcal{I}_{\mathrm{tr}}
    =
    \{i_s,i_s+1,\ldots,i_e\},
    \label{eq:transparent_interval}
\end{equation}
{where $s$ and $e$ denote the indices of the start and end frames, respectively.} in which the output should become fully transparent. We set this as the middle eight frames by default.

At rectified-flow timestep $n$, the clean latent {$\widehat Z_0$} is estimated from the current
latent $Z_n$ and predicted velocity $\widehat v_n$, {which is formulated as}:
\begin{equation}
    \widehat Z_0
    =
    Z_n-n\widehat v_n.
    \label{eq:clean_latent_estimate}
\end{equation}
We decode the frames associated with $\mathcal{I}_{\mathrm{tr}}$:
\begin{equation}
    \widehat V_{\mathcal{I}_{\mathrm{tr}}}
    =
    D_{\mathrm{RGBA}}
    \left(
        \widehat Z_0^{\mathcal{I}_{\mathrm{tr}}}
    \right).
    \label{eq:guided_frame_decode}
\end{equation}
Let $V_{\varnothing}$ denote a fully transparent RGBA target. The
transparent-state objective {$\mathcal{L}_{\mathrm{tr}}$ is formulated as}:
\begin{equation}
    \mathcal{L}_{\mathrm{tr}}
    =
    \frac{1}{|\mathcal{I}_{\mathrm{tr}}|}
    \sum_{i\in\mathcal{I}_{\mathrm{tr}}}
    \left\|
        \widehat V_i-V_{\varnothing}
    \right\|_2^2.
    \label{eq:transparent_state_loss}
\end{equation}
The current video latent is updated through inference-time gradient guidance.
Let $\mathcal{G}$ denote the indices of the guided latent frames. We compute
the gradient only with respect to the corresponding clean latent:
\begin{equation}
    g_n
    =
    \nabla_{\widehat Z_{0,\mathcal{G}}}
    \mathcal{L}_{\mathrm{tr}},
    \bar g_n
    =
    \frac{g_n}{\lVert g_n\rVert_2+\epsilon},
    \widehat Z'_{0,\mathcal{G}}
    =
    \widehat Z_{0,\mathcal{G}}
    -
    \eta_n\bar g_n.
    \label{eq:latent_guidance_update}
\end{equation}
where $g_n$ and $\bar g_n$ denote the raw and normalized guidance gradients,
respectively. The parameter $\eta_n$ controls the guidance step size, while
$\epsilon$ is a small constant for numerical stability. The remaining latent frames are kept unchanged. The updated clean latent
frames are subsequently re-noised to timestep $n$ and merged with the
unchanged frames, yielding the guided noisy latent $Z'_n$ for the next
denoising operation.

An arbitrary RGB keyframe contains complex color, texture, geometry, and
spatial layout, making exact inference-time matching difficult. 
{In contrast, a}
fully
transparent frame
represents a low-complexity state dominated by zero
alpha occupancy. It can therefore be {enforced}
more reliably through gradient
guidance, while the complex appearance of the effect remains controlled by the
trained boundary-conditioning mechanism.


\subsubsection{Inverse Phase Transformation}
\label{sec:inverse_phase}

The guided model produces a phase-reparameterized video following the
positive-zero-positive pattern. We finally apply the inverse cyclic
transformation:
\begin{equation}
    V^{*}
    =
    \mathcal{R}_{k}^{-1}(\widetilde V).
    \label{eq:inverse_rotation}
\end{equation}
This inverse transformation converts the generated sequence back to the target
zero-positive-zero pattern:
\begin{equation}
    +\rightarrow0\rightarrow+
    \quad
    \xrightarrow{\mathcal{R}_{k}^{-1}}
    \quad
    0\rightarrow+\rightarrow0.
    \label{eq:inverse_pattern}
\end{equation}
Temporal phase reparameterization and transparent-state guidance are mutually
dependent. The former converts an interior appearance condition into observable
boundary conditions that can be learned from limited RGBA data, whereas the
latter determines the zero-alpha interval required to recover the original
effect after inverse transformation.

\subsection{{Dataset}}
\label{sec:dataset}

High-quality RGBA videos remain scarce because their alpha channels typically
require professional {content}
creation 
or accurate video matting. We therefore construct
the training data from multiple public image and video matting datasets,
supplemented with RGBA visual effects collected from the Internet. Samples can be found in Figure~\ref{fig:dataset}.

\begin{figure}
    \centering
    \includegraphics[width=0.99\linewidth]{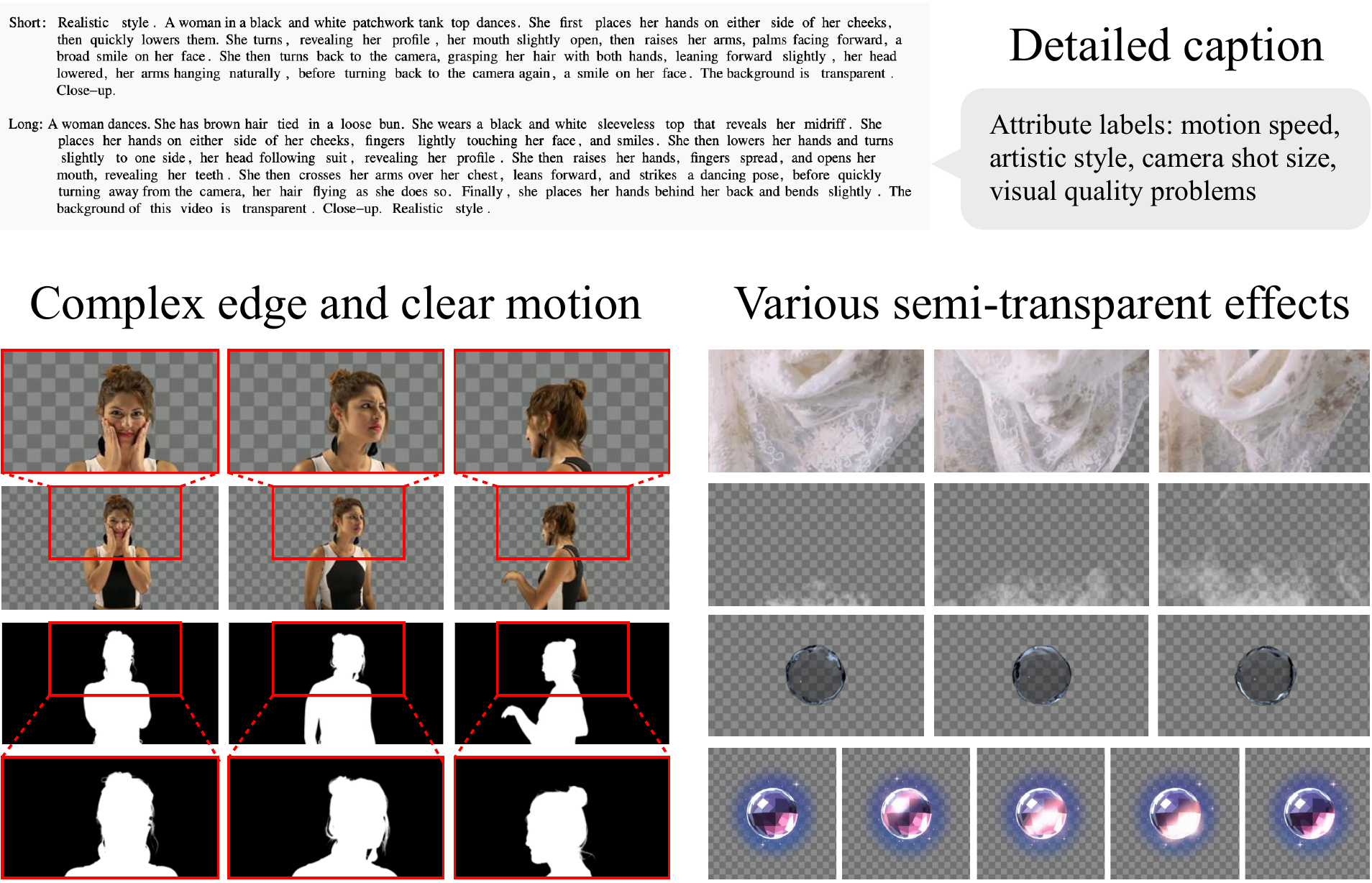}
    \vspace{-2mm}
    \caption{Samples from our dataset. Our videos show clear motion and complex edges, such as hair. Each video has short and long captions, plus attribute labels. The dataset also includes various semi‑transparent effects like sheer fabric, smoke, water, and glow.}
    \label{fig:dataset}
\end{figure}

\noindent{\bf {RGBA VAE Data}~~}
{
For VAE training, we collect images from ten image matting datasets, including
AIM-500~\cite{AIM}, AM-2K~\cite{li2022bridging}, Distinctions-646~\cite{Qiao_2020_CVPR}, HHM2K~\cite{Sun_2023_CVPR}, HHM50K~\cite{Sun_2023_CVPR}, Human2K~\cite{Liu_2021_ICCV}, P3M-10K~\cite{rethink_p3m}, RealWorldPortrait-636~\cite{yu2021mask}, SIMD~\cite{sun2021sim}, and Transparent-460~\cite{cai2022TransMatting}. Video
samples are obtained from DVM~\cite{sun2021dvm}, VideoMatting108~\cite{zhang2021attention}, VideoMatte240K~\cite{lin2021real}, and additional
Internet sources. Image samples are converted into static video clips and
processed using randomized temporal windows so that they can be jointly trained
with video data. We divide the resulting collection into 95\% training and 5\%
validation subsets, yielding 77,237 training clips and 4,066 validation clips.
}

\noindent{\bf {Text-to-RGBA Generation Data}~~}{The diffusion transformer is adapted using a carefully selected subset of the
same collection. We prioritize samples with clear motion, complex alpha
boundaries, semi-transparent content, and lighting effects. For each sample,
Qwen2.5-VL-72B~\cite{qwen2.5-VL} is used to generate both short and
detailed captions. We additionally annotate motion speed, artistic style,
camera shot size, and potential visual-quality issues. The final generation
dataset contains 429 samples, consisting of 301 clips from video matting
datasets, 43 images from image matting datasets, and 85 Internet videos
containing special effects. Although compact, this curated dataset covers
challenging transparency phenomena such as sheer fabric, smoke, water, glow,
and fine-grained hair boundaries.}

\noindent{\bf Image-to-RGBA Generation Data~~}
We first reuse the 429 samples from the
text-to-RGBA generation dataset to adapt the pretrained image-to-video model to
the RGBA domain, {obtaining}
our base image-to-RGBA model. We then collect an
additional set of 196 RGBA effect videos for temporal-pattern adaptation. Each
video follows an emergence-disappearance process, in which the effect gradually
emerges from a fully transparent state, remains visible in the intermediate
frames, and eventually vanishes into transparency. We apply the proposed
temporal phase reparameterization to these videos and use them to fine-tune the
RGBA first-and-last-frame video model. This compact effect dataset allows the
model to learn the target temporal evolution without requiring large-scale
middle-frame-conditioned RGBA training data.
\section{Experiments}
\label{sec:experiments}


\subsection{Experimental Setup}
\label{sec:experimental_setup}

\subsubsection{Implementation Details}
\label{sec:implementation_details}
In Figure~\ref{fig:image_rgba_training_framework}, the purple dotted line means the specific process for image-to-RGBA model. For text-to-RGBA model, the input is the original video without effect and temporal transformation.

\noindent{\bf {Text-to-RGBA Model}~~}
{Our text-to-RGBA video generator is built upon Wan2.1-T2V-14B~\cite{wan2025wan}. Unless otherwise specified, the model generates 81 frames at a resolution of $480\times832$ and 16 FPS. We adopt a four-step accelerated sampling schedule without classifier-free guidance. The LoRA rank of both RGB and alpha VAE decoders is 128, while the DoRA rank of the diffusion transformer is 32. The RGBA VAE is trained for 75K iterations with a batch size of 2, and the diffusion transformer is adapted for 1,750 iterations with a batch size of 8. Although the default evaluation setting is $480\times832$ and 81 frames, the fully convolutional design supports arbitrary aspect ratios, spatial resolutions, and video lengths at 16 FPS.}

As shown in Fig.~\ref{fig:merge}, we employ 3D causal convolutional components of Wan-VAE~\cite{wan2025wan} to construct the feature-merging module. The encoded RGB and alpha features are first concatenated and fed into a causal 3D convolution block for $t$-iteration feature merging, yielding the intermediate representation. Then, we utilize $n$ submodules, which is set as 5, each consisting of a causal residual block and an attention layer to further integrate RGB and alpha features. Finally, a causal 3D convolution predicts the shared RGBA latent $Z$. We directly predict $Z$ without applying a KL-divergence regularizer, as our objective is to adapt a pretrained latent representation rather than learn a new latent space from scratch.

\begin{figure*}[h]
    \centering
    \includegraphics[width=0.99\linewidth]{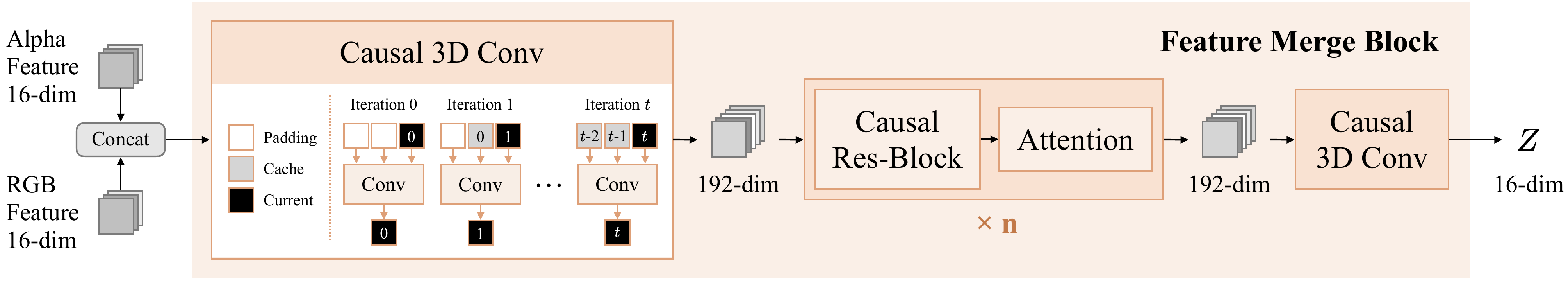}
    \vspace{-2mm}
    \caption{The detailed architecture for the feature merging block. RGB and alpha features are concatenated and fused by a causal 3D convolution. Then, we use several causal residual blocks and attention layers. Finally, a causal 3D convolution generates the merged latent $Z$.}
    \label{fig:merge}
\end{figure*}

\noindent{\bf Image-to-RGBA Model~~}
We adopt the Wan2.1-I2V-14B~\cite{wan2025wan} and the RGBA VAE (see Sec.~\ref{wanvae}) to construct our image-to-RGBA video generation model, which is first trained for $1,070$ iterations. We utilize the trained RGBA I2V model to initialize the phase-reparameterized model and further fine-tune it on $196$ RGBA effect videos for $1,070$ iterations with a batch size of $8$.
During visibility-aware
phase sampling, we set the value of alpha-mass threshold $\tau_m$ defined in
Eq.~\eqref{eq:visible_cut} to 0.5. During inference, transparent-state guidance is applied to the first 32 of the 40
sampling steps, performing 2--3 gradient iterations per step with a normalized step size $\eta_s=50$. All image-conditioned videos consist of 81 frames at a
resolution of $480\times 832$ and are generated using 40 sampling steps.

\noindent{\bf Dataset Split~~}
Our constructed RGBA dataset consists of $77,237$ and $4,066$ video clips for training and validation of the text-to-RGBA video model, respectively. We further curate $429$ samples to train the pretrained image-to-video model, obtaining the base image-to-RGBA video generation model. Then, we additionally collect $196$ RGBA effect videos for temporal-pattern adaptation. We respectively generate $60$ RGBA videos for the evaluation of text-to-RGBA and image-to-RGBA video generation, respectively. No source asset or temporally overlapping clip is shared between the training and testing sets.

\subsubsection{Evaluation Protocol}
\label{sec:evaluation_protocol}

\noindent{\bf Text-to-RGBA Video Generation~~}
We use VBench~\cite{huang2023vbench} to
evaluate aesthetic quality, motion smoothness, and temporal flickering.
Text alignment and visual naturalness are assessed using GPT-4o~\cite{gpt_4o}. Because these evaluators accept only RGB videos, each RGBA
video is composited onto a white background before evaluation. We additionally
conduct a {perceptive evaluation study}
for transparency correctness and overall generation
quality. Participants rank the outputs of all compared methods, and a lower
average rank indicates better performance.

\noindent{\bf Image-to-RGBA Video Generation~~}
We evaluate the image-conditioned video generation models from three complementary perspectives:First, \emph{reference fidelity} measures how well the generated visible frames preserve the input asset. Following DreamBooth~\cite{ruiz2023dreambooth}, we report masked DINO similarity
(\mbox{DINO-I})~\cite{caron2021emerging} and masked CLIP image similarity
(\mbox{CLIP-I})~\cite{radford2021learning} between the input asset and generated
visible frames.
Second, we evaluate general video quality using the same VBench metrics adopted for text-to-RGBA video generation, including aesthetic quality,
motion smoothness, and temporal flickering. We compute DINO-I, CLIP-I, and aesthetic quality on frames whose mean alpha over the non-transparent support is greater than 0.5, as appearance-dependent metrics are strongly affected by fully transparent or nearly empty frames;
Third, we conduct a perceptive evaluation study to assess temporal-effect correctness and overall video quality.

\subsubsection{Compared Methods}
\label{sec:compared_methods}

\noindent{\bf Text-to-RGBA Baselines~~}
We compare against LayerFlow~\cite{ji2025layerflow}, the combination of
LayerDiffuse~\cite{LayerDiffuse} and AnimateDiff~\cite{AnimateDiff}, and both the open-source and closed-source versions of TransPixeler~\cite{wang2025transpixeler}. For LayerFlow, we evaluate its
foreground-layer generation capability. The closed TransPixeler model is
accessed through Adobe Firefly. We additionally compare with a two-stage
solution that combines Wan~\cite{wan2025wan} and MatAnyone~\cite{MatAnyone}.

\noindent{\bf Image-to-RGBA Baselines~~}
As no end-to-end image-to-RGBA video generation models exist, we construct two-stage baselines by integrating Wan-I2V~\cite{wan2025wan} with MatAnyone~\cite{MatAnyone} for normal RGBA video generation, and Wan2.1-FLF2V~\cite{wan2025wan} with MatAnyone~\cite{MatAnyone} for RGBA effect video generation. Specifically, the input RGBA asset is first composited into an RGB image, animated using Wan-I2V or  Wan2.1-FLF2V, and then processed by MatAnyone, with the input alpha channel provided as the initial mask.



\subsection{{Text-to-RGBA Video Generation}}
\label{sec:text_rgba_results}

\noindent{\bf Comparison with Existing Methods}
Table~\ref{tab:text_rgba_quantitative} and
Fig.~\ref{fig:text_rgba_comparison} present the quantitative and qualitative
comparisons, respectively. LayerDiffuse combined with AnimateDiff exhibits
limited motion and weak text alignment; for example, it fails to generate the
prompted transparent glass in the third case. This indicates that directly
transferring an image-level RGBA representation to a video model is insufficient. LayerFlow better exploits pretrained video priors but produces
noticeable object deformation. TransPixeler yields stronger overall results,
yet its generated RGB content remains less realistic in challenging fantasy
scenes and may contain structural artifacts such as malformed hands.
Furthermore, both its open- and closed-source variants produce incorrect glass
transparency in Fig.~\ref{fig:text_rgba_comparison}, revealing unstable
RGB-alpha relationships. In contrast, our method preserves fine structures
such as hair, produces realistic fire and smoke, and correctly represents both
transparent and opaque glass. These visual results are consistent with
Table~\ref{tab:text_rgba_quantitative}: by preserving the pretrained RGB
distribution and shifting the alpha distribution, our method achieves the best
performance across all reported metrics.

\begin{figure*}
    \centering
    \includegraphics[width=0.99\linewidth]{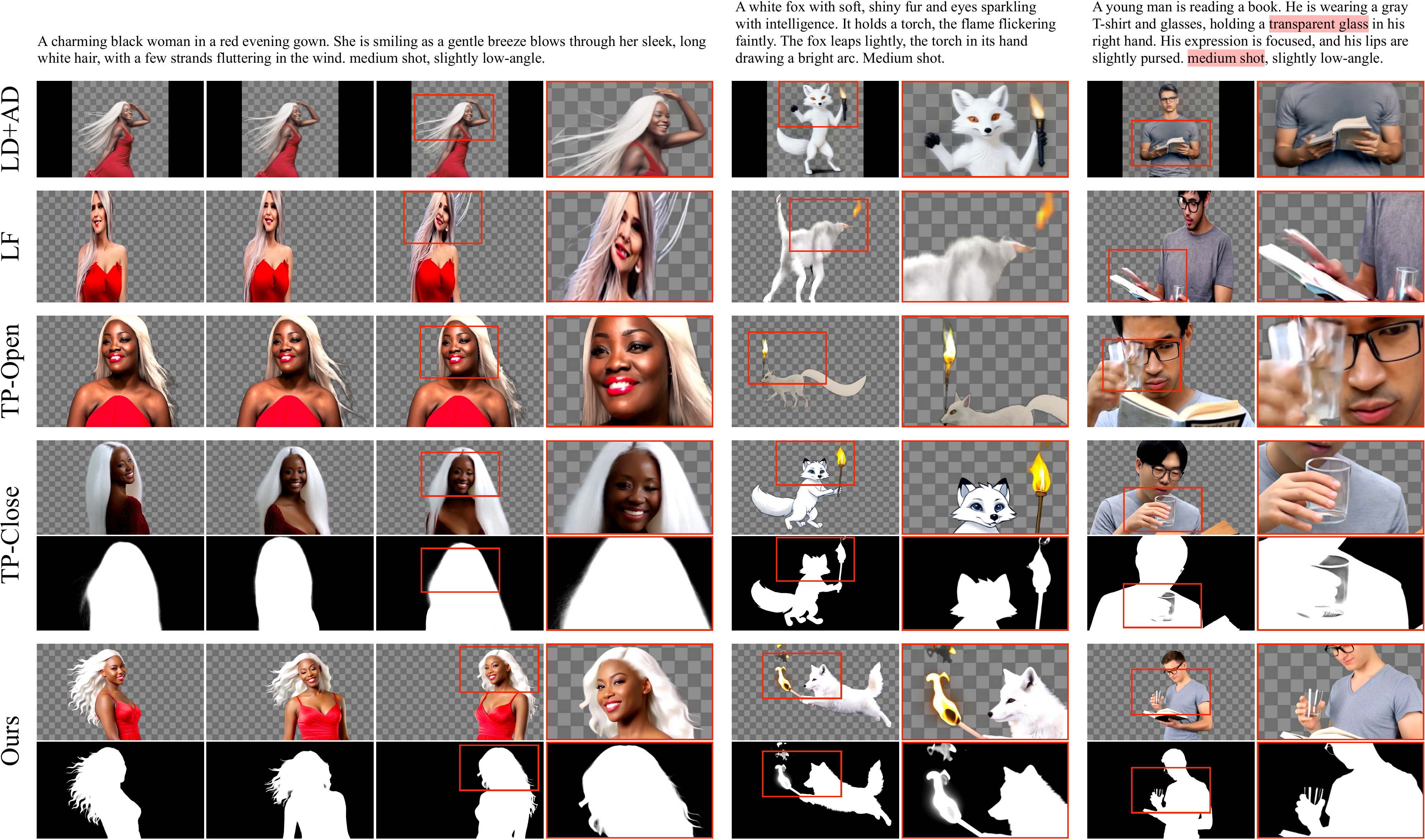}
    \vspace{-2mm}
    \caption{Qualitative comparison with existing text-to-RGBA video generation
        methods. Our method produces more realistic RGB content, finer alpha
        boundaries, and more accurate transparency for challenging objects and
        semi-transparent effects. Please refer to the supplementary material
        for video results.}
    \label{fig:text_rgba_comparison}
\end{figure*}

\setlength{\tabcolsep}{8pt}
\renewcommand{\arraystretch}{1.2}
\begin{table*}[t]
    \centering
    \caption{Quantitative comparison with existing text-to-RGBA video generation
        methods. Our Wan-Alpha consistently achieves superior video generation performance across all five evaluation metrics.
        }
    \vspace{-2mm}
    \footnotesize
    \begin{tabular}{c||ccccc}
    \hline
    \multicolumn{1}{c||}{\multirow{2}*{\makecell[c]{\textbf{Method}}}}  
    & \multirow{2}{*}{\makecell[c]{\textbf{Text} \\ \textbf{Alignment} $\uparrow$}}
    & \multirow{2}{*}{\makecell[c]{\textbf{Aesthetic} \\ \textbf{Quality} $\uparrow$}}
    & \multirow{2}{*}{\bf {Naturalness$\uparrow$}}
    & \multirow{2}{*}{\makecell[c]{\textbf{Motion} \\ \textbf{Smoothness} $\uparrow$}}
    & \multirow{2}{*}{\makecell[c]{\textbf{Temporal} \\ \textbf{Flickering} $\uparrow$}}\\
      & & & & &  \\ \hline\hline
     \textbf{LayerFlow (Single)}~\cite{ji2025layerflow}&2.67&0.535&2.35&0.9837 &0.9788\\ \hline\textbf{LayerDiffuse}~\cite{LayerDiffuse}\textbf{ + AnimateDiff}~\cite{AnimateDiff}&3.15&0.617&3.03&0.9893 &0.9853\\ \hline
     
     \textbf{TransPixeler (Open)~\cite{wang2025transpixeler}}&3.16&0.570&2.97& 0.9821 & 0.9872\\ \hline
     \textbf{TransPixeler (Close)~\cite{wang2025transpixeler}}&3.45&0.573&3.07& 0.9907 & 0.9822\\ \hline
     \textbf{Ours}&\bf{4.00}&\bf{0.649}&\bf{3.19}&\bf{0.9949}&\bf{0.9941} \\ \hline
    \end{tabular}
    \label{tab:text_rgba_quantitative}
\end{table*}

The {results of perceptive evaluation study} 
in Table~\ref{tab:text_rgba_user_study} lead to the same
conclusion. We conduct online user study to evaluate the transparency correctness and overall quality of generated RGBA videos. We randomly order test cases and show them to 40 participants together with results generated by all other methods. Our method obtains the best average ranks for both transparency
correctness and overall quality. Qualitatively, it better preserves fine
structures such as hair and produces more natural semi-transparent effects,
including smoke, fire, illumination, and glass.

\setlength{\tabcolsep}{3pt}
\renewcommand{\arraystretch}{1.2}
\begin{table}[t]
    \centering
    \caption{User ranking for text-to-RGBA generation. Lower transparency and overall indicate better performance.
    }
    \vspace{-2mm}
    \footnotesize
    \begin{tabular}{c||ccccc}
    \hline
    \multicolumn{1}{c||}{\textbf{Method}}    
    & \multirow{1}{*}{\bf {Transparency$\downarrow$}}
    & \multirow{1}{*}{\bf {Overall$\downarrow$}} \\
    \hline\hline
     \textbf{LayerFlow (Single)}~\cite{ji2025layerflow} & 4.29 &3.57 \\ \hline
     \textbf{LayerDiffuse}~\cite{LayerDiffuse}\textbf{ + AnimateDiff}~\cite{AnimateDiff} & 3.40 &4.23 \\ \hline
     \textbf{TransPixeler (Open)}~\cite{wang2025transpixeler} & 2.51 &2.71  \\ \hline
     \textbf{TransPixeler (Close)}~\cite{wang2025transpixeler} & 2.57 &3.37 \\ \hline
     \textbf{Ours} & \textbf{1.23} &\textbf{1.11} \\ \hline
    \end{tabular}
    \label{tab:text_rgba_user_study}
\end{table}

We further compare our method with a two-stage pipeline combining Wan and
MatAnyone, which first generates an RGB video with Wan and then extracts the foreground alpha
channel using MatAnyone. Because MatAnyone requires a user-provided mask for
initialization, the pipeline is not fully automatic. We prompt Wan to generate
the subject against a white background to simplify subsequent matting.
Nevertheless, as shown in Fig.~\ref{fig:matanyone_comparison}, portions of the
white background remain in the resulting RGBA video, leading to inaccurate
alpha boundaries and visible artifacts after compositing. More importantly,
post-hoc matting assumes a separable foreground and background and therefore
cannot reliably recover intrinsically semi-transparent content such as
illumination, smoke, and glow. By jointly generating RGB and alpha, our method
produces cleaner transparent backgrounds, more accurate fine boundaries, and
more faithful semi-transparent effects without requiring an additional mask or
matting stage.
\begin{figure*}
    \centering
    \includegraphics[width=0.99\linewidth]{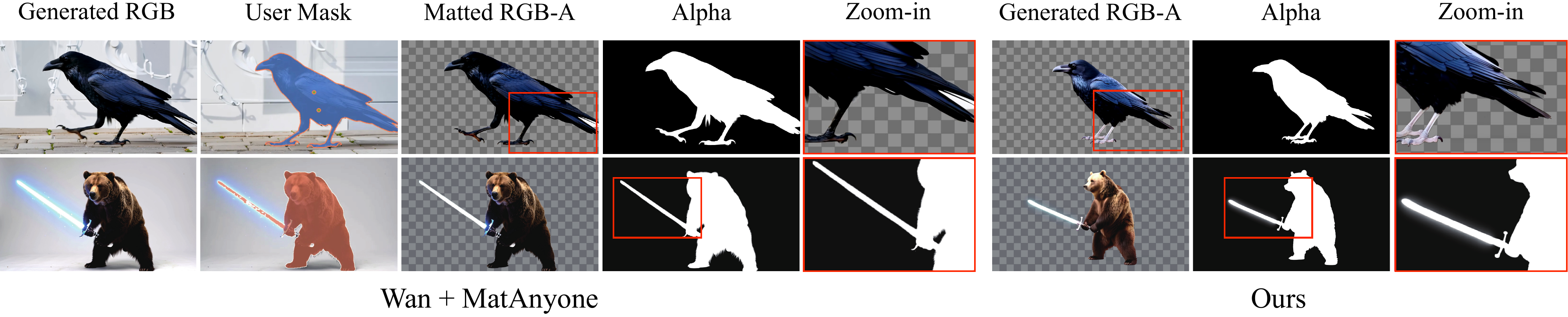}
    \vspace{-3mm}
    \caption{Qualitative comparison with the two-stage Wan + MatAnyone pipeline.
        Despite using a white-background prompt and a user-provided initial
        mask, the matting pipeline leaves residual background content and
        struggles with semi-transparent effects. Our end-to-end RGBA generation
        model produces cleaner alpha channels and more accurate transparency.}
    \label{fig:matanyone_comparison}
\end{figure*}

\noindent{\bf Ablation Study of RGBA Representation Learning}
We evaluate VAE reconstruction quality using PSNR, SSIM~\cite{wang2004image}, and LPIPS~\cite{zhang2018unreasonable}, latent-distribution separation using RGB-Dist and Alpha-Dist, and text-to-RGBA video generation by disabling various components of RGBA representation learning. Specifically, RGB-Dist and Alpha-Dist measure the Euclidean distances (Dist)~\cite{dokmanic2015euclidean} between the adapted and pretrained representations for RGB and alpha content, respectively. An effective representation should minimize RGB-Dist while maximizing Alpha-Dist, preserving the pretrained RGB distribution while shifting the alpha distribution.
As shown in Table~\ref{tab:ablation}, incorporating rendering-based preprocessing and supervision (Rendering) brings significant performance gain in both RGB and alpha reconstruction (see the second row of Table~\ref{tab:ablation}), but failing to preserve the pretrained RGB distribution. We further enable transparency-aware bidirectional diffusion (TABD), yielding enhance performance across all evaluation metrics. Our Wan-Alpha incorporates Rendering, TABD, and transparency-guided mean sampler (MS), which further improves text alignment, aesthetics, motion smoothness, and temporal stability without affecting VAE reconstruction quality.
\setlength{\tabcolsep}{1.2pt}
\renewcommand{\arraystretch}{1.2}
\begin{table*}[h]
    \centering
    
    \caption{Ablation study on RGB-A video reconstruction and generation performance. Our full model achieves the best performance across all metrics, demonstrating the effectiveness of the proposed framework design.}
    \vspace{-0.05in}
    \footnotesize
    \begin{tabular}{cc|c||cccccccc||ccccc}
    \hline
    \multicolumn{3}{c||}{\textbf{Method}}
    &\multicolumn{8}{c||}{\textbf{RGB-A VAE}}
    &\multicolumn{5}{c}{\textbf{Text-to-Video Generation}}
    
    \\ \hline
    \multicolumn{2}{c|}{\textbf{VAE}}
    &\multicolumn{1}{c||}{\textbf{T2V}}
    & \multicolumn{2}{c|}{\textbf{PSNR}$\uparrow$}
    & \multicolumn{2}{c|}{\textbf{SSIM}$\uparrow$}
    & \multicolumn{2}{c|}{\textbf{LPIPS}$\downarrow$}
    & \multicolumn{2}{c||}{\textbf{Dist}}
    & \multirow{2}{*}{\makecell[c]{\textbf{Text} \\ \textbf{Alignment} $\uparrow$}}
    & \multirow{2}{*}{\makecell[c]{\textbf{Aesthetic} \\ \textbf{Quality} $\uparrow$}}
    & \multirow{2}{*}{\bf {\makecell[c]{Naturalness$\uparrow$}}}
    & \multirow{2}*{\makecell[c]{\textbf{Motion} \\ \textbf{Smoothness} $\uparrow$}}
    & \multirow{2}*{\makecell[c]{\textbf{Temporal} \\ \textbf{Flickering} $\uparrow$}}
    \\ \cline{1-11}
    {\bf {\makecell[c]{Rendering}}}
    & \multicolumn{1}{c|}{\bf {\makecell[c]{TABD}}}
    & {\bf {\makecell[c]{MS}}}&\textbf{RGB}&\multicolumn{1}{c|}{\textbf {$\alpha$}}&\textbf{RGB}&\multicolumn{1}{c|}{\textbf {$\alpha$}}&\textbf{RGB}&\multicolumn{1}{c|}{\textbf {$\alpha$}}&\textbf{RGB}$\downarrow$&\multicolumn{1}{c||}{\textbf {$\alpha$}$\uparrow$}&&&&
                    \\ \hline\hline
      & ~ & ~ & 40.12 & \multicolumn{1}{c|}{39.98} & 0.97 & \multicolumn{1}{c|}{0.97} & 0.043 & \multicolumn{1}{c|}{0.025} &0.985&3.56&3.67&0.616&3.17&0.990&0.987 \\ \hline
     \ding{51} &  & ~& 40.88 & \multicolumn{1}{c|}{41.22} & 0.97 & \multicolumn{1}{c|}{0.98} & 0.040 & \multicolumn{1}{c|}{0.023}  &1.015&4.10&3.77&0.619&3.17&0.992&0.984\\ \hline
     \ding{51} & \ding{51}& & \textbf{41.47} & \multicolumn{1}{c|}{\textbf{42.22}} & \textbf{0.97} &\multicolumn{1}{c|}{\textbf{0.98}} & \textbf{0.037} & \multicolumn{1}{c|}{\textbf{0.022}} & \textbf{0.857} &\textbf{4.29}&3.86&0.621&3.19&0.992&0.990 \\ \hline
     \ding{51} & \ding{51}& \ding{51}& \textbf{41.47} & \multicolumn{1}{c|}{\textbf{42.22}} & \textbf{0.97} &\multicolumn{1}{c|}{\textbf{0.98}} & \textbf{0.037} & \multicolumn{1}{c|}{\textbf{0.022}} & \textbf{0.857} &\textbf{4.29}&\textbf{4.00}&\textbf{0.649}&\textbf{3.19}&\textbf{0.995}&\textbf{0.994} \\ \hline
    \end{tabular}
    \label{tab:ablation}
\end{table*}

We observe that opaque regions occasionally exhibit holes, as RGB and alpha remain entangled in the shared latent space in the absence of bidirectional diffusion.
Our proposed objectives, TABD and MS, separate their roles from the perspective of the frozen DiT, resulting in cleaner foreground alpha maps. Fig.~\ref{fig:text_rgba_ablation} qualitatively demonstrates the effectiveness
of TABD and MS. Without both components, the model confuses opaque RGB content with transparency, tending to exhibit
holes and incomplete alpha regions within the foreground. Introducing TABD substantially improves the foreground alpha
structure by making RGB and alpha representations more distinguishable to the DiT. However, the model may still preserve undesired
background content because the shifted transparent latents are more challenging to recover
from standard Gaussian noise. The full model addresses this issue using the transparency-guided mean sampler, producing a cleaner transparent background while preserving foreground appearance and fine alpha boundaries.

\begin{figure}[t]
    \centering
    \includegraphics[width=\linewidth]{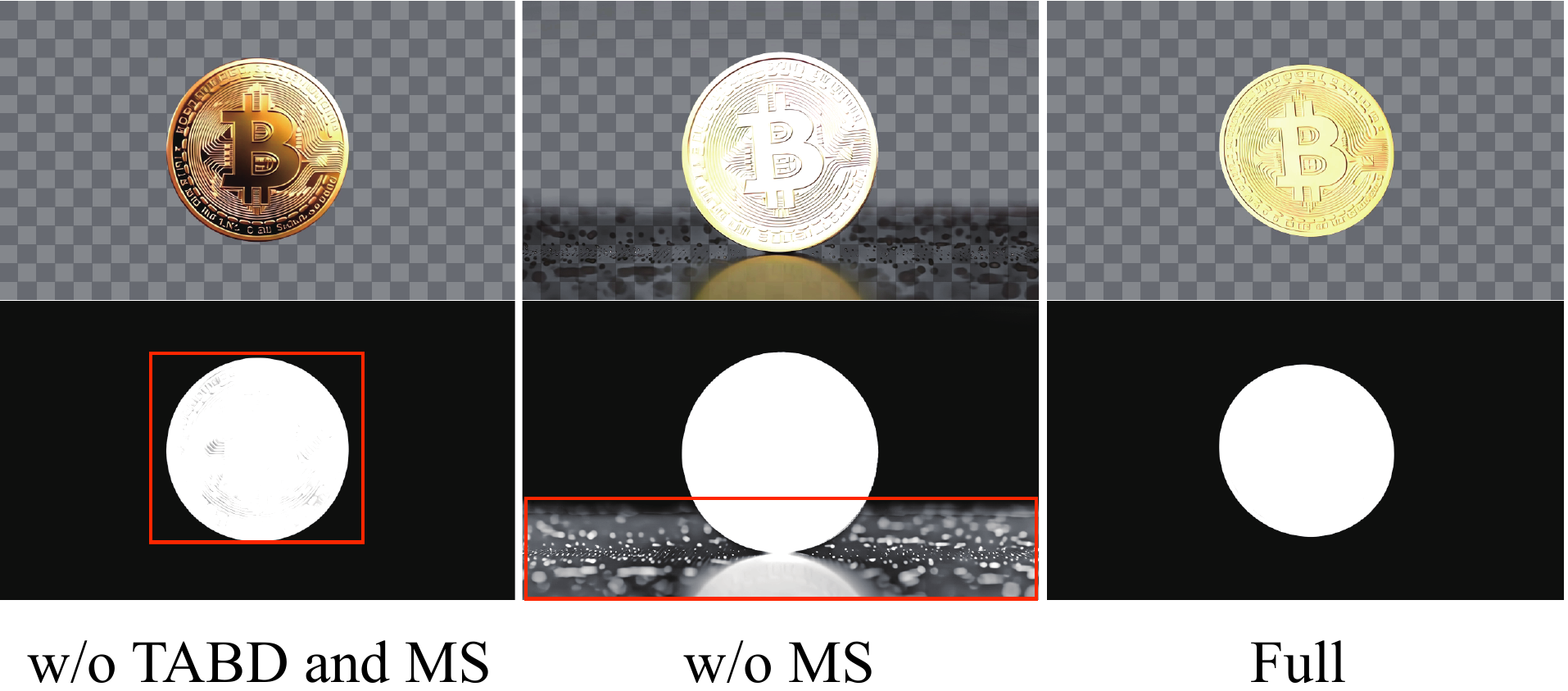}
    \vspace{-2mm}
    \caption{
        {Qualitative ablation of the text-to-RGBA generation framework. TABD
        improves RGB--alpha separation, while MS facilitates clean transparent
        backgrounds.}
    }
    \label{fig:text_rgba_ablation}
\end{figure}
\noindent{{\bf Ablation Study of Transparency-Guided Sampling}}
{Without the transparency-guided mean sampler, the model cannot explicitly
control the placement of generated content and often leaves unwanted background
artifacts. Mean-shifted noise provides a coarse transparency prior while
retaining sufficient flexibility for the model to synthesize fine structures
and motion. We use $\mu=0.05$ by default. A smaller value, such as
$\mu=0.03$, provides insufficient transparency control, whereas an excessively
large value, such as $\mu=0.5$, may introduce a slight color bias.
Fig.~\ref{fig:mean_ablation} provides a qualitative comparison. With
$\mu=0.03$, the shifted noise provides insufficient transparency guidance and
may leave residual background content. The default value $\mu=0.05$ achieves a
favorable balance between transparency control and RGB quality. Increasing the
shift to $\mu=0.5$ excessively perturbs the noise distribution and introduces
visible color bias. Moving or rescaling the Gaussian ellipse correspondingly
changes the approximate position and size of the generated foreground.}
\begin{figure}[t]
    \centering
    \includegraphics[width=\linewidth]{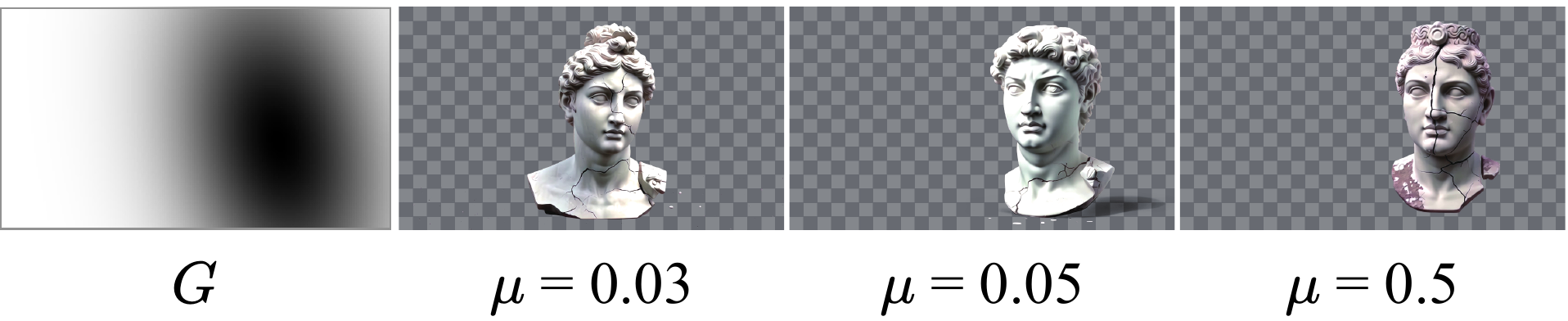}
    \vspace{-2mm}
    \caption{
        Effect of the mean-shift strength $\mu$ in the transparency-guided mean
        sampler.
    }
    \label{fig:mean_ablation}
\end{figure}

\noindent{\bf Spatial Transparency Control~~}The Gaussian ellipse mask provides spatial control over the approximate
position and extent of generated content. Translating the mask moves the
foreground correspondingly, while changing its scale alters the generated
object size. Because the mask specifies only a coarse spatial prior, the model
retains sufficient flexibility to synthesize fine boundaries, motion, and
content-dependent orientation. 
As visualized in
Fig.~\ref{fig:position}, moving the mask from left to right produces a
corresponding displacement of the foreground, demonstrating that the generated
alpha support follows the user-specified spatial layout. Enlarging the mask
increases the foreground extent rather than merely stretching its appearance.
Meanwhile, the model automatically adjusts object pose and orientation to
maintain a natural composition under different mask configurations. These
results show that the Gaussian mask controls coarse transparency geometry
without over-constraining fine structures or temporal dynamics. Additional
qualitative examples confirm this behavior across human subjects, animals,
smoke, fire, and illumination effects.

\begin{figure}[t]
    \centering
    \includegraphics[width=0.99\linewidth]{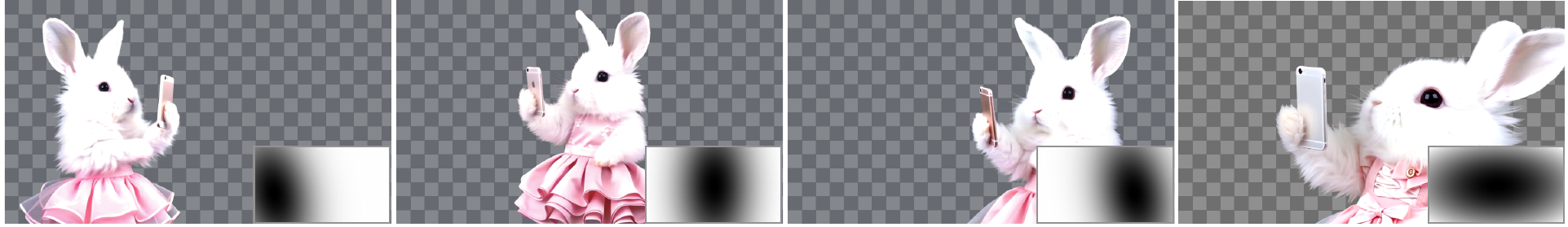}
    \vspace{-2mm}
    \caption{Results under different Gaussian ellipse masks $G$ (shown at the bottom right). Users can use $G$ to control the shape and position of transparent regions.}
    \label{fig:position}
\end{figure}


\subsection{Image-to-RGBA Video Generation}
Image-to-RGBA video generation remains largely unexplored. To the best of our
knowledge, no existing method directly animates a user-provided RGBA asset while
jointly generating its RGB appearance and temporally coherent alpha channel. Therefore, we construct two-stage baselines by integrating Wan2.1-I2V with MatAnyone for normal first-frame-conditioned RGBA video generation, and Wan2.1-FLF2V with MatAnyone for emergence--disappearance RGBA effect video generation, where arbitrary-frame conditioning enables alpha propagation from designated visible reference frames. 
The latter is more challenging because the foreground becomes fully transparent during the sequence, requiring the model to synthesize a complete transition between visible and transparent states while preserving the reference appearance. For both normal and RGBA effect video generation, we compare our direct RGBA video generation models against the corresponding two-stage generation baselines in terms of \textit{reference fidelity} and overall video generation quality.


\noindent{\bf Comparison with Competing Normal RGBA Video Generation Baselines~~}
As reported in Table~\ref{tab:image_rgba_comparison}, our method consistently outperforms the two-stage generation-and-matting pipelines across all metrics.
For normal RGBA videos, our method improves DINO-I from $0.9033$ to $0.9066$ and CLIP-I from $0.9089$ to $0.9445$, demonstrating better preservation of the structure, semantics, and appearance of the input asset. It also achieves slightly higher aesthetic quality, motion smoothness, and temporal flickering scores. Since both methods are built upon Wan2.1-I2V and thus
share similar
RGB generation and motion priors, these consistent improvements indicate that directly modeling RGB and alpha jointly can mitigate
the information loss and propagation errors introduced by post-hoc video matting.

We present qualitative comparisons for normal RGBA video generation in Fig.~\ref{fig:matanyone_comparison_i2v}. Two-stage baseline, Wan+MatAnyone, first uses Wan2.1-I2V to render the asset into an RGB video, where transparency information is irreversibly mixed with the composited background. MatAnyone then relies on an interactively specified segmentation mask to initialize the matting process. This two-stage pipeline inevitably introduces and accumulates errors, particularly around thin structures and low-opacity boundaries. These errors are subsequently propagated to the estimated alpha channel, resulting in overly opaque regions or the loss of soft transparency gradients. In contrast, our proposed model achieves superior \textit{reference fidelity} by jointly modeling RGB appearance and alpha boundaries throughout the video generation process. 
More fundamentally, post-hoc matting cannot reliably recover intrinsic transparency once semi-transparent foreground colors have been blended with the background in the generated RGB frames. 
It therefore tends to produce nearly opaque alpha values or lose soft transparency gradients. 
Beyond these quality limitations, the two-stage pipeline requires separate RGB generation, interactive mask specification, and video matting, making the overall process cumbersome and labor-intensive for practical applications.
In contrast, our method directly generates RGB and alpha channels from the input RGBA asset, eliminating the need for additional user interaction.


\begin{figure*}
    \centering
    \includegraphics[width=0.99\linewidth]{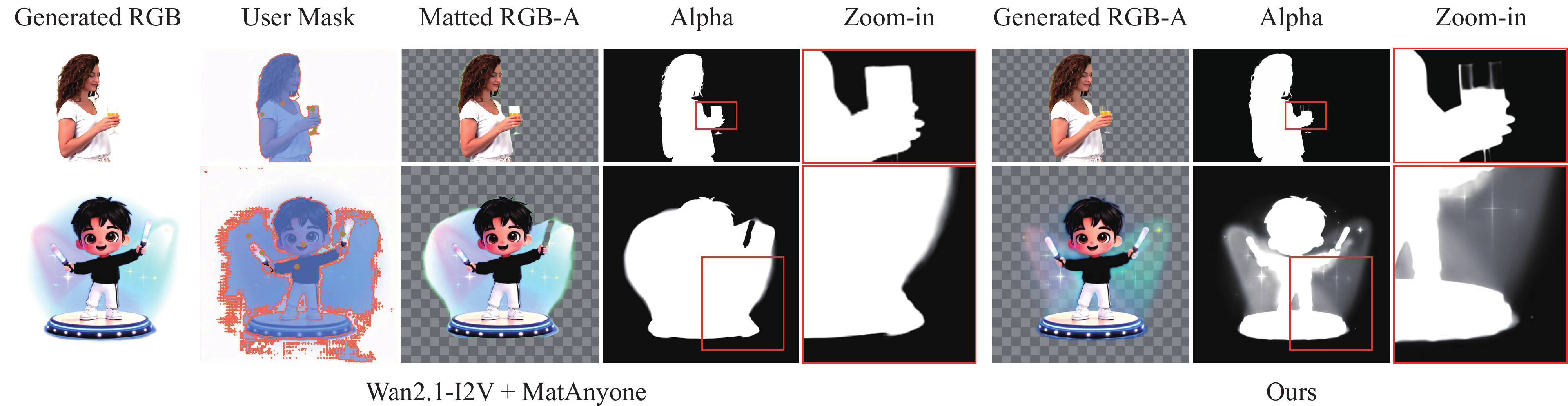}
    \vspace{-2mm}
    \caption{Qualitative comparison with the two-stage Wan2.1-I2V + MatAnyone pipeline.
        Despite using a white-background prompt and a user-provided initial
        mask, the matting pipeline leaves residual background content and
        struggles with semi-transparent effects. Our end-to-end RGBA generation
        model produces cleaner alpha channels and more accurate transparency.}
    \label{fig:matanyone_comparison_i2v}
\end{figure*}

\noindent{\bf Comparison with RGBA Effect Video Generation Baselines~~}
We present quantitative comparisons with the competing baseline in Table~\ref{tab:image_rgba_comparison} and qualitative comparisons in Fig.~\ref{fig:matanyone_comparison_flf2v}. Our proposed framework achieves significant improvements in both DINO-I and CLIP-I over the two-stage Wan2.1-FLF2V+MatAnyone baseline. These performance gains demonstrate that temporal phase reparameterization effectively utilizes the visible conditioning frame to anchor both the structural identity and semantic appearance of the generated effect. Furthermore, our framework consistently achieves superior overall video quality in terms of aesthetic quality, motion smoothness, and temporal flickering. This demonstrates that explicitly learning the transition between visible and transparent states produces more natural content and more temporally coherent alpha evolution than propagating an interactively specified mask after RGB video generation.

Fig.~\ref{fig:matanyone_comparison_flf2v} further reveals the limitations of applying post-hoc video matting to emergence--disappearance effects. This limitation becomes more pronounced for emergence--disappearance RGBA effect video generation. Although MatAnyone supports arbitrary-frame mask conditioning, it primarily propagates foreground masks and does not explicitly model the transition between positive- and zero-alpha states. Therefore, it tends to leave residual alpha during disappearance or produce temporally unstable boundaries during reappearance. 

We investigate two possible initialization positions for the two-stage Wan2.1-FLF2V+MatAnyone pipeline. When MatAnyone is initialized from the first frame (\emph{First}), its interactive mask can provide a complete foreground at the initial frame. However, requiring the first frame to remain visible is inherently conflicted with the desired emergence-disappearance RGBA effect, in which the content is expected to emerge from a transparent state. Therefore, this setting cannot produce the complete transparent-visible-transparent results. Moreover, errors introduced during RGB video generation (see distorted text shown in Fig.~\ref{fig:matanyone_comparison_flf2v}), are propagated to the subsequent matting stage and cannot be effectively corrected afterward.

An alternative is to initialize MatAnyone from a visible middle frame (\emph{Middle}). Although the interactive mask successfully extracts the object
at the condition frame, MatAnyone performs forward propagation from this initialization frame. Therefore, the frames preceding the middle condition are discarded, yielding only an incomplete second half of the RGBA video. Moreover, when the object subsequently disappears from the RGB frames, the propagated matting memory continues to predict a foreground region, resulting in a persistent object-shaped artifact in the alpha channel. This limitation highlights that tracking-oriented video matting does not explicitly model transitions to a fully transparent state.

In contrast, our method treats the visible condition frame as an appearance
anchor 
while explicitly controlling the transparent intervals through temporal phase reparameterization and transparent-state guidance. This design enables our proposed framework to generate the complete
emergence--disappearance process, preserve the conditioned appearance,
and produce clean zero-alpha frames after the effect vanishes.

\begin{figure*}
    \centering
    \includegraphics[width=0.99\linewidth]{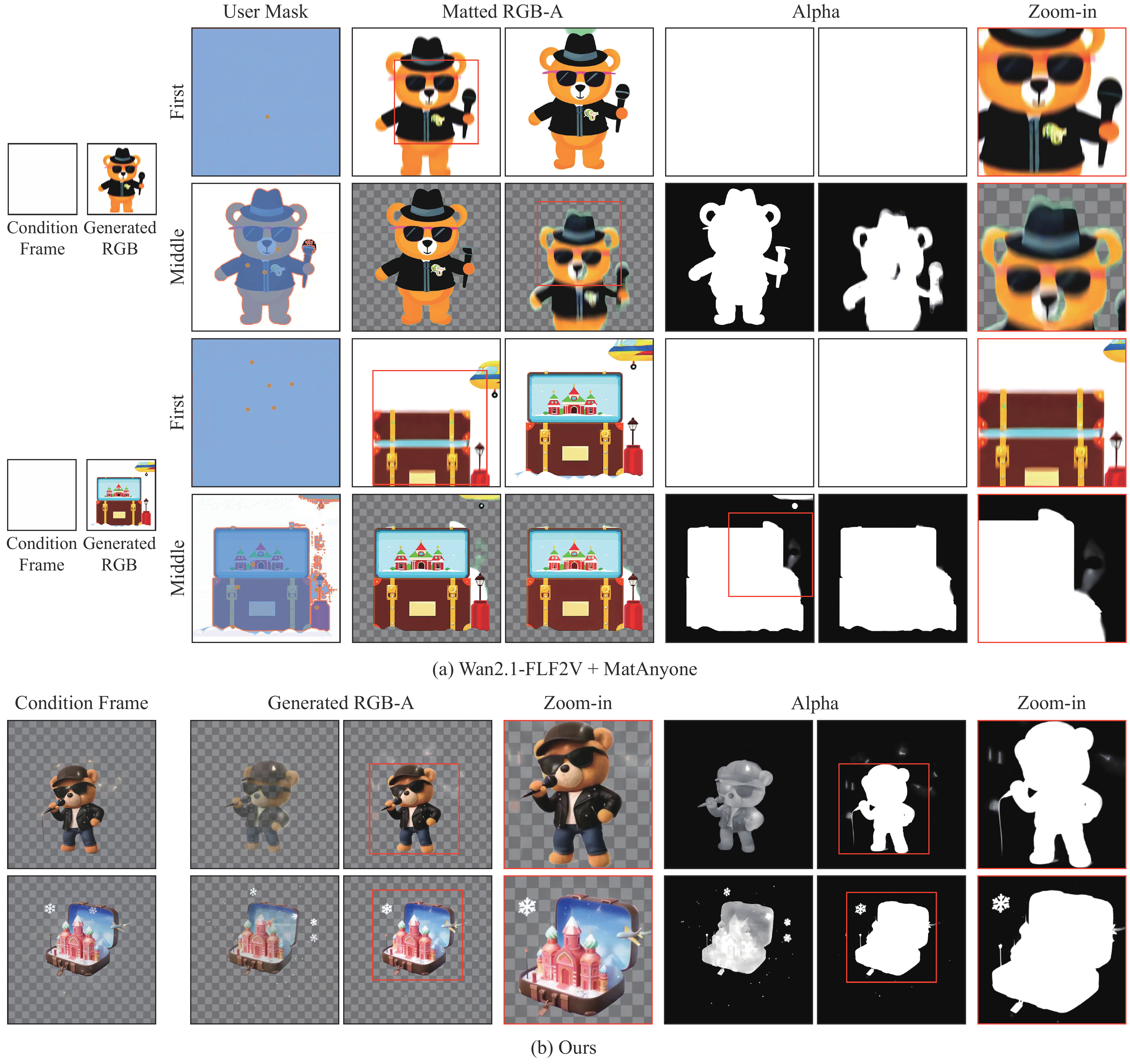}
    \vspace{-2mm}
    \caption{Qualitative comparison with the two-stage Wan2.1-FLF2V + MatAnyone pipeline
under different matting initialization positions. With \emph{First}
initialization, {MatAnyone extracts the visible first frame, but fails to produce the emergence-disappearance RGBA effect.}
The RGB generator also fails to preserve fine appearance details such as text. 
With
\emph{Middle} initialization, MatAnyone only processes the frames following the condition frame and discards the first half of the video. Moreover,
it continues to propagate an object-shaped alpha artifact even after the RGB
content has disappeared. In contrast, our method conditions on a user-specified
visible frame and generates a complete RGBA effect that gradually emerges and
then disappears with clean transparent states.}
    \label{fig:matanyone_comparison_flf2v}
\end{figure*}

\setlength{\tabcolsep}{3.4pt}
\renewcommand{\arraystretch}{1.2}
\begin{table*}[t]
\tiny
    \centering

    \caption{Quantitative comparison on normal and effect image-to-RGBA video
    generation. DINO-I and CLIP-I measure reference fidelity, while the
    remaining metrics evaluate general video quality. 
    }
    \vspace{-2mm}
    \footnotesize
    \begin{tabular}{c|c||ccccc}
    \hline
    \multicolumn{2}{c||}{\textbf{Method}}
    & \multirow{1}{*}{\bf DINO-I$\uparrow$}
    & \multirow{1}{*}{\bf CLIP-I$\uparrow$}
    & \multirow{1}{*}{\bf Aesthetic Quality$\uparrow$}
    & \multirow{1}{*}{\bf Motion Smoothness$\uparrow$}
    & \multirow{1}{*}{\bf Temporal Flickering$\uparrow$}\\ \hline\hline

    \multirow{2}{*}{\textbf{Normal}}
    & \textbf{Wan2.1-I2V + MatAnyone}~\cite{MatAnyone}
    & 0.9033 & 0.9089 & 0.6296 & 0.9912 & 0.9858\\ \cline{2-7}
    & \textbf{Ours}
    & \textbf{0.9066} & \textbf{0.9445}
&\textbf{0.6305} & \textbf{0.9916} & \textbf{0.9864}\\ \hline\hline

    \multirow{2}{*}{\textbf{Effect}}
    & \textbf{Wan2.1-FLF2V + MatAnyone}~\cite{MatAnyone}
    & 0.6679 & 0.7969 & 0.5749 & 0.9917 & 0.9926\\ \cline{2-7}
    & \textbf{Ours}
    & \textbf{0.9227} & \textbf{0.9368} & \textbf{0.5968} & \textbf{0.9941} & \textbf{0.9930}\\ \hline
    \end{tabular}
    \label{tab:image_rgba_comparison}
\end{table*}

\noindent{\bf Perceptive Evaluation Study~~}
We conduct a perceptive evaluation study using randomly ordered test cases. 40 participants are shown the input RGBA asset together with videos generated by all methods and are asked to rank them according to temporal-effect correctness and overall video quality. Our method achieves better results across all metrics.

\setlength{\tabcolsep}{1pt}
\renewcommand{\arraystretch}{1.2}
\begin{table}[t]
\tiny
    \centering
    \caption{User ranking for image-to-RGBA generation. 
    }
    \vspace{-2mm}
    \footnotesize
    \begin{tabular}{c|c||cc}
    \hline
    \multicolumn{2}{c||}{\textbf{Method}}
    & \multirow{1}{*}{\bf Temporal Effect$\downarrow$}
    & \multirow{1}{*}{\bf Overall$\downarrow$}\\ \hline\hline

    \multirow{2}{*}{\textbf{Normal}}
    & \textbf{Wan2.1-I2V + MatAnyone}
    & 1.73 & 1.62 \\ \cline{2-4}
    & \textbf{Ours}
    & \textbf{1.27} & \textbf{1.38} \\ \hline\hline

    \multirow{2}{*}{\textbf{Effect}}
    & \textbf{Wan2.1-FLF2V + MatAnyone}
    & 1.88 & 1.84 \\ \cline{2-4}
    & \textbf{Ours}
    & \textbf{1.12} & \textbf{1.16} \\ \hline
    \end{tabular}
    \label{tab:image_rgba_user}
\end{table}

\noindent{\bf Ablation Study~~}
Our image-to-RGBA effect generation framework contains two principal designs: temporal phase reparameterization and transparent-state guidance. We examine the effect of removing each component, with the results reported in Table~\ref{tab:image_rgba_component_ablation}. 
We first remove both components, degrading the model into the Wan2.1-FLF2V-based RGBA video generation model that uses
effect videos in their original temporal order. 
Since the first and last frames of these videos are fully transparent, the boundary conditions contain neither observable RGB appearance nor spatial alpha structure. Consequently, the image-conditioning mechanism becomes ineffective and degrades into text-to-RGBA video generation, resulting in low similarity between the input asset and generated visible video frames (see the first row of Table~\ref{tab:image_rgba_component_ablation}). 

We then incorporate temporal phase reparameterization, which transforms the original zero-positive-zero sequence into a positive-zero-positive sequence. The resulting visible boundary frames provide effective appearance conditions, substantially improving \textit{reference fidelity} (see the second row of Table~\ref{tab:image_rgba_component_ablation}). However, temporal phase reparameterization alone
only enables the model to learn the distribution of
disappearance-emergence RGBA video effects. With limited training data, this learned distribution is insufficient to reliably determine the location and duration of the transparent interval during inference. Finally, we apply transparent-state guidance during inference to explicitly constrain the user-specified interval toward transparency. Combined with temporal phase reparameterization, the full Wan-Alpha++ preserves the conditioned appearance while generating a stable and controllable emergence-disappearance process, achieving the best overall performance.

\begin{table*}[t]
    \centering
    \caption{Component ablation of temporal phase reparameterization (\textbf{Reparameterization}) and
    transparent-state guidance (\textbf{Guidance}). We report reference fidelity, general video quality, and
the user ranking for temporal-effect correctness. Lower is better for Temporal
Effect, while higher is better for all other metrics.}
    \vspace{-2mm}\label{tab:image_rgba_component_ablation}
    \setlength{\tabcolsep}{8pt}
    \renewcommand{\arraystretch}{1.2}
    \begin{tabular}{cc||cccccc}
        \hline
        \multirow{2}*{\makecell[c]{\textbf{Reparameterization}}}
        & \multirow{2}*{\makecell[c]{\textbf{Guidance}}}
        & \multirow{2}{*}{\makecell[c]{\textbf{DINO-I}} $\uparrow$}
        & \multirow{2}{*}{\makecell[c]{\textbf{CLIP-I}} $\uparrow$}
        & \multirow{2}{*}{\makecell[c]{\textbf{Aesthetic} \\ \textbf{Quality} $\uparrow$}}
        & \multirow{2}*{\makecell[c]{\textbf{Motion} \\ \textbf{Smoothness} $\uparrow$}}
        & \multirow{2}*{\makecell[c]{\textbf{Temporal} \\ \textbf{Flickering} $\uparrow$}}
        & \multirow{2}*{\makecell[c]{\textbf{Temporal} \\ \textbf{Effect} $\downarrow$}}
        \\
       & & &&&&&\\ \hline\hline
         & &
0.6315 & 0.7542 & 0.5793 & 0.9919 & 0.9918 & 2.76 \\ \hline
\ding{51} & &
0.9136 & 0.9275 & 0.5924 & 0.9936 & 0.9927 & 2.09 \\ \hline
\ding{51} & \ding{51}
& \textbf{0.9227}
& \textbf{0.9368}
& \textbf{0.5968}
& \textbf{0.9941}
& \textbf{0.9930}
& \textbf{1.12} \\ \hline
    \end{tabular}
\end{table*}

\noindent{\bf Sensitivity to Guidance Hyperparameters~~}
We change the guidance step size and the number of guided sampling steps to examine their effects on video generation performance, with the results reported in Table~\ref{tab:guidance_sensitivity}. A larger step size (e.g., $\eta_s=50$) provides stronger control over the transparent state, leading to consistent improvements across all metrics. However, an excessively large step size (e.g., $\eta_s=75$) overly perturbs the generated appearance, resulting in degraded performance.
For the guidance duration, applying the guidance step for $16$ or $24$ slightly improves \textit{reference fidelity} in terms of DINO-I and CLIP-I, but results in less stable temporal RGBA video effects. Extending guidance to the first $32$ of $40$ sampling steps achieves the best overall video generation performance across aesthetic quality, motion smoothness, and temporal consistency. However, applying guidance throughout all $40$ steps introduces excessive cumulative latent updates, degrading details in visible frames. Therefore, we adopt $\eta_s=50$ and $32$ guided sampling steps as the default configuration.

\begin{table*}[t]
    \centering
    \caption{Sensitivity to transparent-state guidance hyperparameters.}
    \vspace{-2mm}
    \label{tab:guidance_sensitivity}
    \setlength{\tabcolsep}{6pt}
    \renewcommand{\arraystretch}{1.2}
    \begin{tabular}{lcccccc}
        \hline
        \textbf{Parameter} & \textbf{Value}
        & \textbf{DINO-I}$\uparrow$
        & \textbf{CLIP-I}$\uparrow$
        & \multirow{1}{*}{\bf Aesthetic Quality$\uparrow$}
    & \multirow{1}{*}{\bf Motion Smoothness$\uparrow$}
    & \multirow{1}{*}{\bf Temporal Flickering$\uparrow$}
        
         \\
        \hline\hline
       \multirow{4}{*}{Step size $\eta_s$}
& 10
& 0.9208 & 0.9335 & 0.5908 & 0.9935 & 0.9925 \\
& 25
& 0.9220 & 0.9352 & 0.5947 & 0.9939 & 0.9928 \\
& \textbf{50}
& \textbf{0.9227} & \textbf{0.9368} & \textbf{0.5968}
& \textbf{0.9941} & \textbf{0.9930} \\
& 75
& 0.9188 & 0.9310 & 0.5921 & 0.9934 & 0.9924 \\ \hline
\multirow{4}{*}{Guided sampling steps}
& 16
& \textbf{0.9235} & \textbf{0.9372} & 0.5905 & 0.9934 & 0.9924 \\
& 24
& 0.9230 & 0.9369 & 0.5943 & 0.9938 & 0.9927 \\
& \underline{32}
& 0.9227 & 0.9368 & \textbf{0.5968}
& \textbf{0.9941} & \textbf{0.9930} \\
& 40
& 0.9175 & 0.9295 & 0.5897 & 0.9931 & 0.9920 \\ \hline
    \end{tabular}
\end{table*}

\section{Conclusion}

In this work, we present Wan-Alpha++, a unified framework for high-quality
and controllable RGBA video generation. We first address the entanglement
between RGB appearance and alpha transparency by learning shiftable RGB--alpha
distributions in both latent and noise spaces. The proposed
transparency-aware bidirectional diffusion objective preserves the pretrained
RGB distribution while making alpha representations more distinguishable, and
the transparency-guided mean sampler improves transparency synthesis while
providing coarse spatial control over the generated content.
We further extend RGBA video generation from text conditioning to practical
image conditioning. For emergence-disappearance effects, fully transparent
boundary frames invalidate conventional image-to-video conditioning. Temporal
phase reparameterization addresses this problem by moving visible interior
content to the temporal boundaries, allowing the appearance-conditioning prior
of a pretrained FLF2V model to be reused with limited RGBA data.
Transparent-state guidance subsequently controls the location and duration of
the invisible interval, and inverse phase transformation restores the desired
emergence-disappearance process. Together, these designs provide complementary
spatial and temporal control over RGBA generation.
Experiments demonstrate that Wan-Alpha++ improves RGB quality, reference
fidelity, transparency rendering, motion smoothness, and temporal consistency
over existing RGBA generation and two-stage video-matting pipelines. The
generated assets can be directly composited onto arbitrary backgrounds,
supporting applications in visual effects, filmmaking, game development, and
digital content creation. Future work may extend the proposed temporal
reparameterization to more complex multi-stage transparency events and reduce
the additional inference cost introduced by gradient guidance.

\section*{Acknowledgements}
This work was supported by the National Natural Science Foundation of China (No.62476192). It was also funded by Laboratory for Artificial Intelligence in Design (Project 3.1), the InnoHK initiative of the Innovation and Technology Commission of the Hong Kong Special Administrative Region Government.

\bibliographystyle{ieeetr}
\bibliography{wanalpha}
\vspace{-0.4in}

\end{document}